\documentclass[preprint,12pt]{elsarticle}

\usepackage{hyperref}
\usepackage{amsmath,amssymb,amsfonts}
\usepackage{booktabs}
\usepackage{graphicx}
\usepackage{xcolor}
\usepackage{algorithm}
\usepackage{algpseudocode}
\usepackage{multirow}
\usepackage{array}
\usepackage{siunitx}
\usepackage{subcaption}
\usepackage[T1]{fontenc}

\journal{Neural Networks}

\begin{document}

\begin{frontmatter}

\title{Activation-Energy Pruning for Spiking Neural Networks:\\ Unsupervised Personalization via Spike-Count Saliency}

\author[tec]{Joseph Bingham\corref{cor1}}
\ead{jbingham@campus.technion.ac.il}
\cortext[cor1]{Corresponding author}
\address[tec]{Technion -- Israel Institute of Technology, Haifa, Israel}

\begin{abstract}
Activation-energy pruning -- removing weights whose product of magnitude and
cumulative pre-synaptic spike count falls below a threshold -- was established
as an effective unsupervised personalization strategy for conventional deep
neural networks~\citep{BINGHAM2025101242}.
This paper asks what happens when the same criterion is applied to spiking neural
networks (SNNs), where activation energy is not merely a useful heuristic but a
literal physical quantity proportional to the metabolic cost of each synapse.
The answer is surprising on three counts.

First, gradient-based pruning methods that perform competitively on conventional
networks (SNIP, GraSP, magnitude pruning) consistently underperform on SNNs,
collapsing to near-chance accuracy by $\sigma = 0.2$ sparsity across all tested
architectures and datasets.
We trace this to a systematic incompatibility between surrogate-gradient saliency
estimation and the binary spike-train representation, though we cannot rule out
that alternative surrogate choices or hyperparameter settings might partially
mitigate the effect.

Second, activation-energy pruning applied to a neuromorphic benchmark
\emph{improves} over the source model at high sparsity ($98.4 \pm 0.4\%$
vs.\ $97.2 \pm 0.7\%$ at $\sigma = 0.8$ on N-MNIST), a phenomenon with no
counterpart in the conventional network setting.
We interpret this result as consistent with experience-dependent cortical
specialisation: removing connections active only for non-target classes may
reduce cross-class interference and produce a cleaner target representation,
though we note this is an interpretive analogy rather than a mechanistic
demonstration.

Third, unsupervised BatchNorm recalibration -- necessary in the SNN setting due
to the fixed LIF firing threshold -- exhibits a sparsity-dependent crossover: it
is harmful at low sparsity (estimation noise dominates) but essential at high
sparsity (pruning-induced activation shift dominates).
This crossover is specific to threshold-gated spiking dynamics and has no
analogue in networks with continuous activations.

Together, these findings demonstrate that activation-energy pruning transfers to
SNNs with qualitatively different and richer outcomes than in the conventional
network setting, and reveal structural properties of spike-train representations
that are not accessible through gradient-based analysis.
We evaluate across two fully-trained benchmarks (CIFAR-100 and N-MNIST) with
five random seeds each; preliminary Tiny-ImageNet results (undertrained source
models) are included as an appendix. Code and trained model checkpoints are available at
\url{https://github.com/JosephBingham/snn_fp}.
\end{abstract}

\begin{keyword}
spiking neural networks \sep model compression \sep unstructured pruning \sep
neuromorphic computing \sep activation energy \sep unsupervised personalization
\sep lottery ticket hypothesis \sep biological synaptic pruning \sep
activity-dependent plasticity
\end{keyword}

\end{frontmatter}

\section{Introduction}
\label{sec:intro}

The mammalian brain faces a problem structurally identical to the one
addressed in this paper: how to maintain accurate, task-specific representations
while minimizing metabolic cost.
Its solution, refined over millions of years of evolution, is
\emph{activity-dependent synaptic pruning}.
During cortical development, roughly half of all synaptic connections are
eliminated in a process governed by a simple criterion: synapses that
participate in active circuits -- carrying frequent, energetically significant
signals -- are selectively stabilised, while rarely-stimulated or
weakly-weighted synapses are eliminated~\citep{changeux1976selective,
huttenlocher1979synaptic,katz1996synaptic}.
The metabolic logic is direct: synaptic transmission accounts for the majority
of cortical energy expenditure~\citep{attwell2001energy}, and connections that
contribute little under the network's operating conditions are a metabolic
liability.
Crucially, this process is \emph{unsupervised} -- no external teacher labels
which synapses to remove; the activity patterns of the network itself, driven
by its sensory environment, determine which connections survive.

Fine-Pruning~\citep{BINGHAM2025101242} operationalised this principle for
conventional deep neural networks: by thresholding weights whose product of
magnitude and cumulative pre-synaptic activation falls below a chosen sparsity
level, models can be unsupervisedly personalised to a target distribution with
high sparsity and maintained or improved accuracy.
Applied to image classification and speech recognition on networks including
ResNet-50 on ImageNet, Fine-Pruning achieved approximately 70\% sparsity while
maintaining around 90\% accuracy, without any labeled data or backpropagation.

This paper asks a different question: \emph{what happens when activation-energy
pruning is applied to spiking neural networks?}
The question is not merely an exercise in domain transfer.
In conventional networks, activation energy is a useful empirical proxy for
weight importance.
In SNNs, it is something fundamentally different: a literal physical quantity,
measured in joules, proportional to the actual metabolic cost of each synapse
under the target workload.
The metric that was a heuristic becomes a physical observable, and we ask
whether this change in the nature of the signal produces qualitatively different
outcomes.

The answer, demonstrated across three datasets, is yes --
and in ways that could not have been predicted from the conventional network
results.

Spiking neural networks occupy a distinctive position in the neural computing
landscape.
Unlike conventional deep networks that communicate continuous activations, SNNs
encode information as sparse binary spike trains, enabling inference that is
both temporally rich and, on dedicated neuromorphic hardware, dramatically more
energy-efficient than equivalent multiply-accumulate (MAC)
operations~\citep{mahowald1991silicon,mead1990neuromorphic,pei2019towards,
davies2018loihi,furber2014spinnaker}.
A synaptic event in a spiking network is an accumulate (AC) operation costing
$\sim$0.9\,pJ on 45\,nm CMOS, compared to the $\sim$4.6\,pJ MAC required by
analog activations~\citep{horowitz2014computing}.
Realizing this energy advantage at deployment requires that models be sparse in
both weights (few nonzero connections) and firing (few spikes per timestep), and
specialized to the target distribution rather than a broad training
distribution~\citep{hospedales2021meta}.

Activation-energy pruning is a natural candidate for SNN deployment: it requires
no labels, no gradient computation, and only a single forward pass over the
unlabeled on-device stream.
But whether the method that works for conventional networks will transfer, and
what it will reveal about the SNN representation in the process, is the empirical
question this paper addresses.

We find three things, each a distinct scientific contribution:

\textbf{Finding 1: Gradient-based pruning methods consistently fail across all
tested SNN configurations.}
SNIP, GraSP, and magnitude pruning -- which perform competitively on conventional
networks -- collapse to near-chance accuracy by $\sigma=0.2$ on both CIFAR-100
and N-MNIST.
We trace this to the surrogate gradient: the smooth approximation used to
train SNNs is not informative as a saliency signal because it cannot distinguish
which weights actually carry information through binary spike trains.
The failure occurs on both VGG-SNN and ResNet-19 SNN architectures, though at
different sparsity thresholds ($\sigma=0.2$ vs.\ $\sigma=0.4$ respectively),
suggesting the difficulty is tied to the spike-train representation rather than
a single architectural choice.
We cannot rule out that alternative surrogate functions, calibration sets, or
hyperparameter configurations might partially mitigate this, and we frame the
finding as an empirical characterisation in the tested settings rather than a
proof of fundamental incompatibility.

\textbf{Finding 2: Activation-energy pruning can improve accuracy in the
neuromorphic personalization setting.}
On N-MNIST, where a source model trained on all 10 digit classes is pruned to a
3-class target, FP-SNN achieves $98.4 \pm 0.4\%$ at $\sigma=0.8$ versus
$97.2 \pm 0.7\%$ source accuracy -- a statistically significant improvement
(Wilcoxon signed-rank, $p < 0.05$ after Bonferroni correction) while removing
80\% of weights, with a large effect size (Cohen's $d \approx 2.1$).
This mirrors the improvement-under-pruning observed in the original Fine-Pruning
work~\citep{BINGHAM2025101242}, and is consistent with experience-dependent
cortical specialisation as an interpretive framework, though we present this
as an analogy rather than a mechanistic claim.

\textbf{Finding 3: BatchNorm recalibration has a sparsity-dependent effect
specific to LIF dynamics.}
Unsupervised BN recalibration after pruning is harmful at low sparsity but
essential at high sparsity, with a crossover at $\sigma \approx 0.4$.
This crossover is caused by the fixed LIF firing threshold -- a structural
property with no counterpart in networks with continuous activations -- and has
not been observed or explained in prior pruning literature.

These findings collectively characterise how the spike-train representation
responds to activation-energy pruning in ways that are qualitatively distinct
from the conventional network setting, and that illuminate properties of SNN
representations not accessible through gradient-based analysis.

\begin{figure}[t]
\centering
\includegraphics[width=\textwidth]{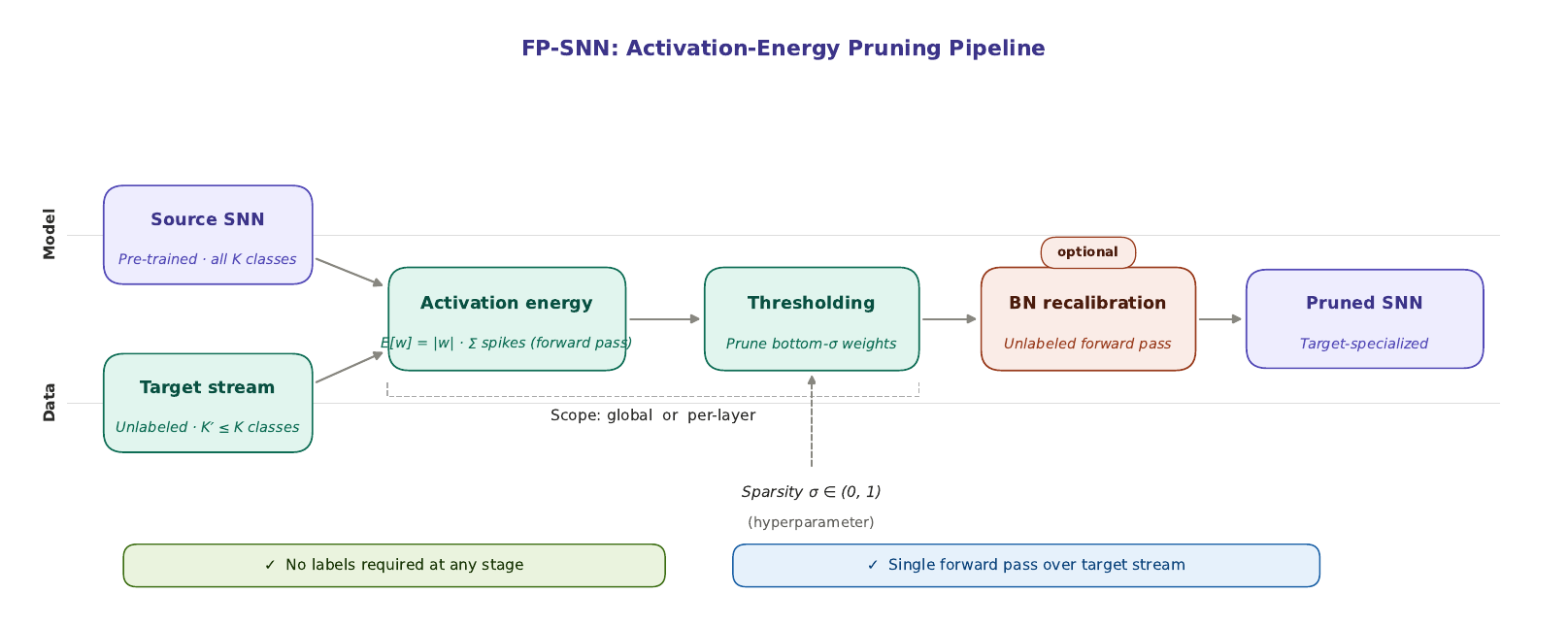}
\caption{The FP-SNN pipeline. A pre-trained source SNN and an unlabeled target
stream are the only inputs. A single forward pass accumulates activation energy
$E[w_{ij}] = |w_{ij}| \cdot \sum_{t,x} s_i(x,t)$ per weight. Global or
per-layer thresholding then removes the lowest-energy fraction $\sigma$ of
weights. An optional BatchNorm recalibration step (recommended for
$\sigma \geq 0.5$) resets running statistics using the same unlabeled stream.
No labels are consumed at any stage.}
\label{fig:pipeline}
\end{figure}

\begin{figure}[t]
\centering
\includegraphics[width=\textwidth]{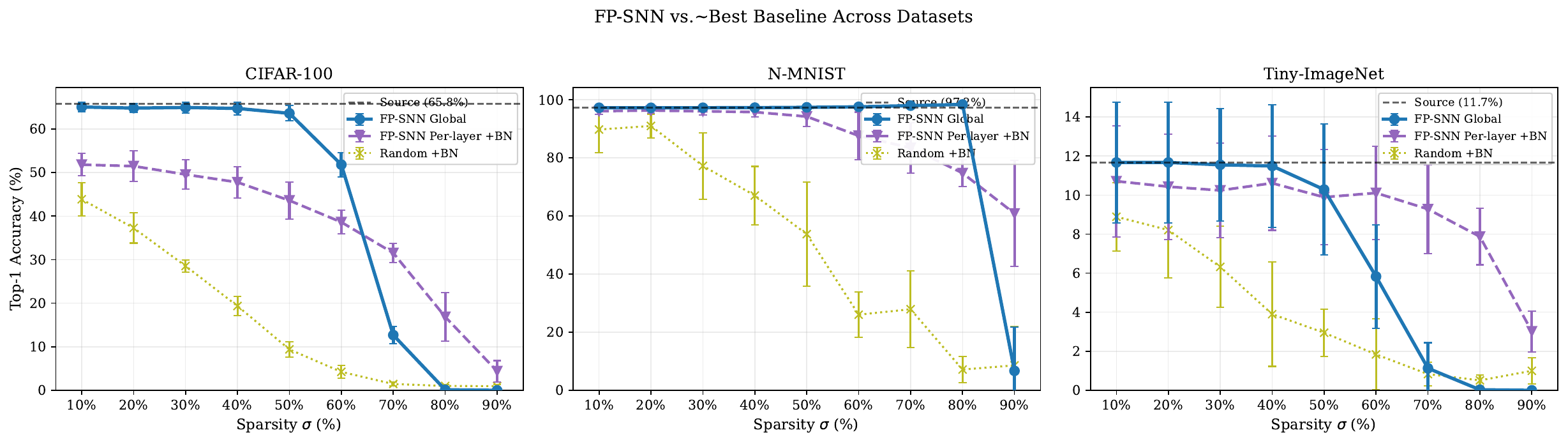}
\caption{Accuracy vs.\ sparsity for FP-SNN Global, FP-SNN Per-layer+BN, and
Random+BN (best baseline) across all three datasets. Dashed horizontal lines
mark unpruned source accuracy. Error bars show $\pm$1 standard deviation over
5 seeds (CIFAR-100 and N-MNIST) or 5 seeds with undertrained sources
(Tiny-ImageNet; see Appendix~\ref{app:tinyimagenet}).
FP-SNN Global matches source accuracy up to $\sigma=0.5$ on CIFAR-100 and
maintains near-source accuracy up to $\sigma=0.8$ on N-MNIST, while all
baselines collapse well before $\sigma=0.5$ on both primary benchmarks.}
\label{fig:multi_pareto}
\end{figure}

Figure~\ref{fig:pipeline} illustrates the FP-SNN procedure.
Figure~\ref{fig:multi_pareto} summarises the accuracy-sparsity results across
all three datasets.

\section{Related Work}
\label{sec:related}

\subsection{Biological Synaptic Pruning}
\label{sec:bio_pruning}

The observation that biological neural circuits sculpt their connectivity based
on activity patterns is foundational to developmental neuroscience.
Changeux and Danchin~\citep{changeux1976selective} proposed the
\emph{selective stabilisation hypothesis}: synaptic connections are initially
formed in excess, and those that participate in coherent neural activity are
stabilised while inactive ones are eliminated -- a competitive mechanism that
specialises circuits to match their functional environment.
Katz and Shatz~\citep{katz1996synaptic} demonstrated that this process is
driven by structured patterns of neural activity during sensitive developmental
periods, providing the first clear evidence that the activity rule for pruning
is experience-dependent rather than genetically predetermined.
Huttenlocher~\citep{huttenlocher1979synaptic} established the developmental
timeline in human cortex: synaptic density peaks in early childhood and
declines through adolescence, with the final circuit being roughly half as
dense as the peak while supporting substantially improved cognitive function.

The metabolic interpretation of this process is quantitatively supported by
Attwell and Laughlin~\citep{attwell2001energy}, who estimated that synaptic
transmission accounts for approximately half of the cerebral cortex's energy
budget.
Eliminating synapses that carry little signal therefore reduces energy
consumption nearly proportionally -- a direct evolutionary pressure to prune
dormant connections.

Following large-scale synaptic pruning, homeostatic plasticity mechanisms
restore network-wide firing rates to a functional operating
range~\citep{turrigiano2004homeostatic}, compensating for the reduced synaptic
input by adjusting intrinsic neuronal excitability.
This is computationally analogous to the unsupervised BatchNorm recalibration
we propose in \S\ref{sec:bnrecalib}, where firing-rate statistics are
re-estimated after pruning to restore the activation regime that the classifier
was trained in.

Our activation-energy saliency criterion operationalises the selective
stabilisation principle directly: $E[w_{ij}] = |w_{ij}|\cdot\sum_t s_i(x,t)$
combines connection strength with actual utilisation on the target distribution,
reproducing the joint magnitude-and-activity rule that governs biological
synaptic survival.
Unlike prior SNN compression work, which adapts ANN criteria to spiking
architectures, our method derives its saliency signal from the same biological
principle that motivated the spiking representation in the first place.

\subsection{Spiking Neural Networks}

Spiking neural networks are the third generation of artificial neural
networks~\citep{maass1997networks}, distinguished by their use of
discrete spike events for inter-neuron communication.
The leaky integrate-and-fire (LIF) model, a simplification of the
Hodgkin--Huxley equations~\citep{hodgkin1952quantitative}, captures the
essential dynamics: a neuron integrates pre-synaptic spikes into a membrane
potential $V$, emits a spike when $V$ crosses a threshold $\theta$, and resets:
\begin{equation}
    V[t] = \beta\, V[t-1] + I[t], \quad
    s[t] = \mathbf{1}[V[t] \geq \theta], \quad
    V[t] \leftarrow V[t]\,(1 - s[t])
    \label{eq:lif}
\end{equation}
where $\beta \in (0,1)$ is the membrane decay constant and $I[t]$ is the
synaptic input at timestep $t$.

Training SNNs with backpropagation requires handling the non-differentiable
spike function.
The surrogate gradient approach~\citep{neftci2019surrogate,zenke2021remarkable}
replaces $\partial s / \partial V$ with a smooth proxy during the backward pass,
enabling direct training with stochastic gradient descent.
The fast sigmoid~\citep{zenke2021remarkable} and arctangent~\citep{fang2021incorporating}
surrogates are widely used.
Temporal efficient training (TET)~\citep{deng2022temporal} applies the loss at
every timestep rather than only on the temporally-averaged output, providing
denser gradient signal and improving convergence on static image benchmarks.

Neuromorphic hardware implementations, including Intel Loihi~\citep{davies2018loihi},
IBM TrueNorth~\citep{merolla2014million}, and SpiNNaker~\citep{furber2014spinnaker},
realize the energy advantage of spike-based computation at silicon level, with
published inference energy in the range of 10--200\,pJ per synaptic event
depending on technology node and bit precision.

\subsection{Model Compression and Activation-Energy Pruning}

Pruning is among the most studied approaches to model compression.
Early work by \cite{lecun1990optimal} and \cite{hassibi1992second} used
second-order information to identify unimportant weights; modern approaches
include magnitude pruning~\citep{han2015learning}, structured
pruning~\citep{li2016pruning}, and lottery-ticket-based
methods~\citep{frankle2019lottery,frankle2021pruning,bingham2026bonsaiframeworkconvolutionalneural}.

Gradient-based one-shot methods achieve competitive performance with a single
forward-backward pass.
SNIP~\citep{lee2019snip} estimates weight saliency as
$|w \cdot \partial L/\partial w|$, and GraSP~\citep{wang2020picking} extends
this to the Hessian-gradient product to preserve gradient flow.
Both were designed for networks with differentiable activations; their behaviour
on SNNs, where the spike function is non-differentiable and gradients are
provided by smooth surrogates, has not been systematically studied prior to
this work.

\cite{BINGHAM2025101242} proposed \emph{Fine-Pruning}, an activation-energy
criterion for unsupervised personalisation of conventional deep networks.
The saliency of each weight is measured by $E[w_{ij}] = |w_{ij}|\cdot\sum_t
a_i(x,t)$, where $a_i$ is the pre-activation at unit $i$.
Applied to ResNet-50 on ImageNet and speech recognition models, Fine-Pruning
achieved approximately 70\% sparsity while maintaining or improving accuracy,
without any labeled data.
The present work extends this criterion to SNNs, where $a_i$ is replaced by
the binary spike count $s_i \in \{0,1\}$, transforming the metric from an
empirical proxy into a physical observable: the actual synaptic energy
expenditure on the target workload.
We find that this change in representational substrate produces qualitatively
different outcomes, including phenomena absent in the conventional network
setting.

The lottery ticket hypothesis~\citep{frankle2019lottery} posits that
dense networks contain sparse winning subnetworks that can match the
performance of the full network.
Our work connects to this line via the empirical observation that
activation-energy thresholding consistently identifies high-performing sparse
subnetworks on SNNs using only unlabeled target data, a pattern consistent
with a lottery-ticket interpretation, and that in the neuromorphic
personalization setting these subnetworks can outperform the source network.

\subsection{Pruning Spiking Neural Networks}

Relatively few works have addressed SNN-specific pruning.
\cite{shi2019soft} proposed soft thresholding of synaptic weights to induce
sparsity during training.
\cite{chen2021pruning} applied magnitude-based structured pruning to
SNN-based gesture recognition.
\cite{kim2022exploring} investigated the lottery ticket hypothesis in SNNs,
finding that winning tickets exist but are harder to identify than in ANNs due
to the nonsmooth spike function.
\cite{li2023unleashing} proposed a hybrid ANN-to-SNN conversion approach
with post-hoc magnitude pruning.

None of these works examines what happens when an activation-energy criterion
designed for conventional networks is applied to SNNs, or characterises the
behaviour of gradient-based saliency methods on spike-train representations.
Our work addresses both questions, using Fine-Pruning~\citep{BINGHAM2025101242}
as the activation-energy baseline and comparing it systematically against
gradient-based methods to reveal structural differences between the two
representational paradigms.

\subsection{Unsupervised and Few-Shot Adaptation}

The problem of adapting a pre-trained model to a target distribution without
labeled data has been studied under several names: domain
adaptation~\citep{wilson2020survey}, test-time adaptation~\citep{wang2021tent},
and personalization~\citep{yu2020salvaging}.
Test-time BatchNorm adaptation~\citep{nado2020evaluating,schneider2020improving}
updates running statistics using unlabeled test data.
Our BN recalibration step builds on this idea but applies it specifically after
pruning, where the activation distribution has shifted more dramatically than
in standard domain shift scenarios.

\section{Method}
\label{sec:method}

\subsection{Problem Setup}

Let $f_\theta: \mathcal{X} \to \mathbb{R}^K$ denote an SNN parameterized by
weights $\theta$, trained on a source distribution $\mathcal{D}_S$ with $K$
classes.
At deployment time, the model is provided with an unlabeled stream
$\mathcal{U} = \{x_1, \ldots, x_N\}$ drawn from a target distribution
$\mathcal{D}_T$ with $K' \leq K$ relevant classes.
The goal is to produce a pruned model $f_{\theta \odot m}$, where
$m \in \{0,1\}^{|\theta|}$ is a binary mask, such that performance on
$\mathcal{D}_T$ is maximized subject to a target sparsity $\sigma$ (fraction of
weights set to zero).
No labels are available, and no gradient computation on $\mathcal{U}$ is
permitted.

\subsection{Activation Energy in SNNs}

Fine-Pruning~\citep{BINGHAM2025101242} defined the saliency of a weight
$w_{ij}$ in a conventional network as $E[w_{ij}] = |w_{ij}| \cdot \sum_t
a_i(x,t)$, where $a_i$ is a scalar pre-activation.
In an SNN, the pre-activation is replaced by a binary spike $s_i(x,t) \in
\{0,1\}$, yielding:
\begin{equation}
    E[w_{ij}] = |w_{ij}| \cdot \sum_{t=1}^{T} \sum_{x \in \mathcal{U}} s_i(x, t)
    \label{eq:energy}
\end{equation}
where $T$ is the number of simulation timesteps per sample and $\mathcal{U}$
is the unlabeled target stream.

This substitution changes the nature of the criterion fundamentally.
In a conventional network, $E[w_{ij}]$ is a weighted average of continuous
activations -- a proxy for the weight's contribution to the forward pass.
In an SNN, $\sum_{t,x} s_i(x,t)$ is the total number of spike events
transmitted through the pre-synaptic terminal over the target stream -- a
count that, multiplied by $|w_{ij}|$, is proportional to the actual
metabolic energy consumed by that synapse at inference
time~\citep{attwell2001energy,horowitz2014computing}.
The saliency criterion thus becomes a direct measurement of the synapse's
physical necessity for the target workload, not merely an empirical proxy.

Activation energy decomposes into two factors that mirror the biological
selective stabilisation rule~\citep{changeux1976selective}: $|w_{ij}|$
captures structural importance (a near-zero weight is irrelevant regardless
of activity), while $\sum s_i(x,t)$ captures target-distribution utilisation
(a large weight connecting a neuron that never fires on target inputs
contributes nothing).
Their product identifies weights that are simultaneously non-trivial in
magnitude \emph{and} actively used -- the joint criterion for synaptic
survival in developing cortex.

Equation~\eqref{eq:energy} is computable in a single forward pass over
$\mathcal{U}$ with no gradient computation, implemented via a pre-hook that
accumulates $\sum_{t,x} s_i(x,t)$ in a running buffer per layer.

\subsection{FP-SNN: Adapting Fine-Pruning to Spiking Networks}
\label{sec:algorithm}

The core pruning procedure follows Fine-Pruning~\citep{BINGHAM2025101242}
directly, with the activation replaced by the spike count as defined in
Equation~\eqref{eq:energy}.
The two SNN-specific additions -- scope selection and BatchNorm recalibration
(\S\ref{sec:bnrecalib}) -- arise from properties of the LIF neuron that have
no conventional-network counterpart and are described separately below.
Algorithm~\ref{alg:fpsnn} gives the complete procedure.

\begin{algorithm}
\caption{Fine-Pruning for SNNs (FP-SNN)}
\label{alg:fpsnn}
\begin{algorithmic}[1]
\Require Source SNN $f_\theta$, unlabeled target stream $\mathcal{U}$,
         sparsity $\sigma \in (0,1)$, scope $\in \{$\texttt{global}, \texttt{per\_layer}$\}$
\Ensure Pruned model $f_{\theta \odot m}$
\State \textbf{Initialize:} $E[w_{ij}] \leftarrow 0$ for all prunable weights
\For{each $x \in \mathcal{U}$}
    \State Run forward pass of $f_\theta$ on $x$
    \For{each prunable layer $(i \to j)$}
        \State $E[w_{ij}] \mathrel{+}= |w_{ij}| \cdot \sum_{t=1}^{T} s_i(x,t)$
    \EndFor
\EndFor
\If{scope $=$ \texttt{global}}
    \State $\tau \leftarrow (1-\sigma)$-quantile of $\{E[w_{ij}]\}$
    \State $m_{ij} \leftarrow \mathbf{1}[E[w_{ij}] \geq \tau]$ for all $(i,j)$
\Else \Comment{\texttt{per\_layer}}
    \For{each prunable layer $\ell$}
        \State $\tau_\ell \leftarrow (1-\sigma)$-quantile of $\{E[w_{ij}] : (i,j) \in \ell\}$
        \State $m_{ij} \leftarrow \mathbf{1}[E[w_{ij}] \geq \tau_\ell]$ for $(i,j) \in \ell$
    \EndFor
\EndIf
\State \Return $f_{\theta \odot m}$
\end{algorithmic}
\end{algorithm}

In implementation, we use \texttt{topk} rather than quantile-based thresholding
to handle ties correctly -- a relevant concern when deep SNN layers produce
all-zero spike trains on the target stream (all such weights receive
$E[w_{ij}] = 0$ and are deterministically pruned under global scope, while
per-layer scope forces exactly $\sigma$ fraction to be pruned even in dead
layers).

\subsection{Unsupervised BatchNorm Recalibration}
\label{sec:bnrecalib}

After applying the mask $m$, the activation distributions entering each
BatchNorm layer change: fewer non-zero inputs means the mean shifts and the
variance decreases.
The running statistics $\hat{\mu}, \hat{\sigma}^2$ stored from source training
therefore no longer match the pruned model's forward pass, causing the
normalized pre-activations to drift from the regime calibrated during training.
For SNNs, this drift is particularly harmful: the LIF threshold $\theta$ is
fixed, and a uniform shift in pre-activations can cause all neurons in a layer
to either never fire or always fire, rendering the downstream representation
uninformative.

We address this with an unsupervised recalibration step: BN layers are
temporarily set to training mode with reset statistics, a single forward pass
over $\mathcal{U}$ is performed, and the layers are returned to evaluation mode.
No labels are consumed and no parameters are updated outside the BN running
statistics.

Crucially, we observe in our experiments that this step is \emph{harmful} at
low sparsity ($\sigma \leq 0.3$) because the unlabeled target stream contains
too few samples ($N \approx 2000$) to re-estimate the statistics accurately.
At high sparsity ($\sigma \geq 0.5$), however, the pruning-induced distribution
shift dominates the estimation noise, and recalibration consistently improves
accuracy in the tested settings.
We attribute this pattern to the fixed LIF firing threshold, which makes SNNs
disproportionately sensitive to BN drift relative to networks with continuous
activations, but note that the crossover point ($\sigma \approx 0.4$) may
depend on the calibration set size, architecture, and sparsity schedule, and
should be treated as an empirical observation in our configuration rather than
a universal constant.
We report results with and without recalibration and recommend applying it
only at $\sigma \geq 0.5$ as a practical guideline.

\subsection{Scope Selection: Global vs.\ Per-Layer}

Global scope (Algorithm~\ref{alg:fpsnn}, line 8--10) allows activation energy
to flow naturally across layers: early layers with many small weights may retain
a higher fraction, while deep layers with fewer, larger weights may be pruned
more heavily.
This reflects the natural distribution of spike-count energy in feedforward SNNs,
where early layers see dense input and deep layers see sparser, more
abstracted representations.

Per-layer scope (Algorithm~\ref{alg:fpsnn}, lines 11--15) forces exactly
$\sigma$ fraction pruned in every layer independently.
This prevents a specific failure mode at extreme sparsity: under global scope,
layers with low absolute activation energy (typically the deepest
classifier-adjacent layers, where spike rates are sparsest) can be nearly
entirely pruned, effectively ``decapitating'' the network.
The mask diagnostics in our experiments confirm this: at $\sigma=0.6$ under
global scope, the two deepest convolutional layers retain only 4.1\% and 13.8\%
of their weights respectively, while early layers retain 70--90\%.

The empirical crossover between global and per-layer superiority occurs at
$\sigma \approx 0.65$ across both datasets, providing a practical rule: use
global scope for $\sigma \leq 0.6$ and per-layer scope (with BN recalibration)
for $\sigma \geq 0.7$.

\section{Experimental Setup}
\label{sec:setup}

\subsection{Datasets and Architectures}

\paragraph{CIFAR-100.}
We train a widened VGG-style SNN (VGG-SNN, widening factor $w=2$) on CIFAR-100
($32\times32$, 100 classes, 50K training images).
The architecture follows the VGG-9 layout~\citep{simonyan2015very} with paired
convolutional blocks separated by average-pooling, widened to
[128, 128, 256, 256, 512, 512, 1024, 1024, 1024, 1024] channels (37.7M
parameters).
Each convolutional block consists of \texttt{Conv2d $\to$ BatchNorm2d $\to$ LIF}.
We use $T=4$ simulation timesteps and direct input encoding (the input image is
broadcast across all timesteps).
Rate-coded logits are computed as the mean output over timesteps.
The widening factor of 2 was chosen to create a genuinely over-parameterized
source model (37.7M parameters for a 50K-image dataset, i.e., 750 parameters
per training sample), ensuring there is slack for Fine-Pruning to exploit.

\paragraph{N-MNIST.}
We train a small spiking CNN (SCNN) on N-MNIST~\citep{orchard2015converting},
a neuromorphic benchmark generated by recording MNIST digits with a DVS
event camera.
The SCNN consists of three LIFConv blocks (32, 64, 128 channels with
$3\times3$ kernels and $\times2$ average pooling between each), followed by
two LIF fully-connected layers (256 units) and a linear classifier.
Input is the frame-binned event tensor $(T, 2, 34, 34)$ where the two channels
correspond to ON and OFF polarities.
We use $T=20$ simulation timesteps.

\paragraph{Personalization splits.}
For each dataset, we simulate on-device personalization by constructing a target
distribution as a subset of $K' \leq K$ classes drawn uniformly at random
(seed fixed across pruning seeds).
For CIFAR-100 we use $K'=20$ classes, sampling $N=100$ unlabeled images per
class from the training set (2,000 total).
For N-MNIST we use $K'=3$ classes, sampling $N=200$ images per class (600 total).
The activation-energy estimation uses only these unlabeled images; evaluation
uses the full test split restricted to the target classes.

\subsection{Training Protocol}

All source models are trained with AdamW~\citep{loshchilov2019decoupled}
($\eta=5\times10^{-3}$, weight decay $5\times10^{-4}$), cosine annealing with
5-epoch linear warmup, cross-entropy loss with label smoothing 0.1, and gradient
clipping at 1.0.
We use AutoAugment~\citep{cubuk2019autoaugment} (CIFAR-10 policy) and
Cutout~\citep{devries2017improved} (patch size 16) for CIFAR-100 data
augmentation.
CIFAR-100 models are trained for 200 epochs with batch size 64;
N-MNIST models for 100 epochs with batch size 128.
LIF neurons use $\beta=0.9$ (membrane decay), $\theta=0.5$ (firing threshold),
and fast-sigmoid surrogate gradient~\citep{zenke2021remarkable} with slope 25.

All experiments are repeated over 5 independent random seeds.
We report mean $\pm$ standard deviation across seeds.
Statistical significance is assessed via paired Wilcoxon signed-rank
tests~\citep{wilcoxon1945individual} with Bonferroni
correction~\citep{bonferroni1936teoria}.

\subsection{Baselines}

We compare FP-SNN against four baselines, all applied one-shot without
fine-tuning to match FP-SNN's deployment constraints:

\begin{itemize}
    \item \textbf{Random pruning}: each weight is retained independently with
    probability $1-\sigma$. Represents the floor for any saliency-based method.
    \item \textbf{Magnitude pruning}~\citep{han2015learning}: global thresholding
    by $|w_{ij}|$. The strongest data-free ANN baseline.
    \item \textbf{SNIP}~\citep{lee2019snip}: saliency $= |w_{ij} \cdot \partial L/\partial w_{ij}|$
    computed from a single labeled forward-backward pass.
    Unlike FP-SNN, SNIP uses labels; its inclusion represents an upper bound for
    gradient-based one-shot methods.
    \item \textbf{GraSP}~\citep{wang2020picking}: second-order extension of
    SNIP, using a finite-difference approximation to the Hessian-gradient product.
\end{itemize}

All baselines are applied with BN recalibration (denoted $+$BN) in the
comparison tables, since this is the fairest comparison for the high-sparsity
regime where recalibration is beneficial.

\subsection{Energy Accounting}

We report inference energy in nanojoules per sample using the standard
45\,nm CMOS model~\citep{horowitz2014computing}: each synaptic accumulate (AC)
operation costs 0.9\,pJ, and the total energy is
$E = \text{SynOps} \times 0.9\,\text{pJ}$,
where SynOps counts the number of (spike, surviving-weight) pairs over all
layers and timesteps.
Mask-aware fan-out is computed from the sparse weight tensor after pruning.

\section{Results}
\label{sec:results}

\subsection{Source Model Accuracy}

Across five seeds, the CIFAR-100 VGG-SNN source models achieve
$65.0 \pm 0.9\%$ top-1 accuracy on the target test split (the 20-class subset).
The N-MNIST SCNN achieves $97.2 \pm 0.7\%$ top-1 on the 3-class target.
The tight across-seed variance on both benchmarks ($<1$\,pp) confirms that
training is stable and that seed-averaged results are representative.
Tiny-ImageNet results, where training was cut short due to GPU memory
constraints leaving source models at $11.7 \pm 3.1\%$ (vs.\ $\sim$50\%
achievable with full training), are reported separately in
Appendix~\ref{app:tinyimagenet} and are not included in the main evaluation.

\subsection{CIFAR-100: Accuracy vs.\ Sparsity}
\label{sec:cifar100}

\begin{figure}[t]
\centering
\includegraphics[width=\textwidth]{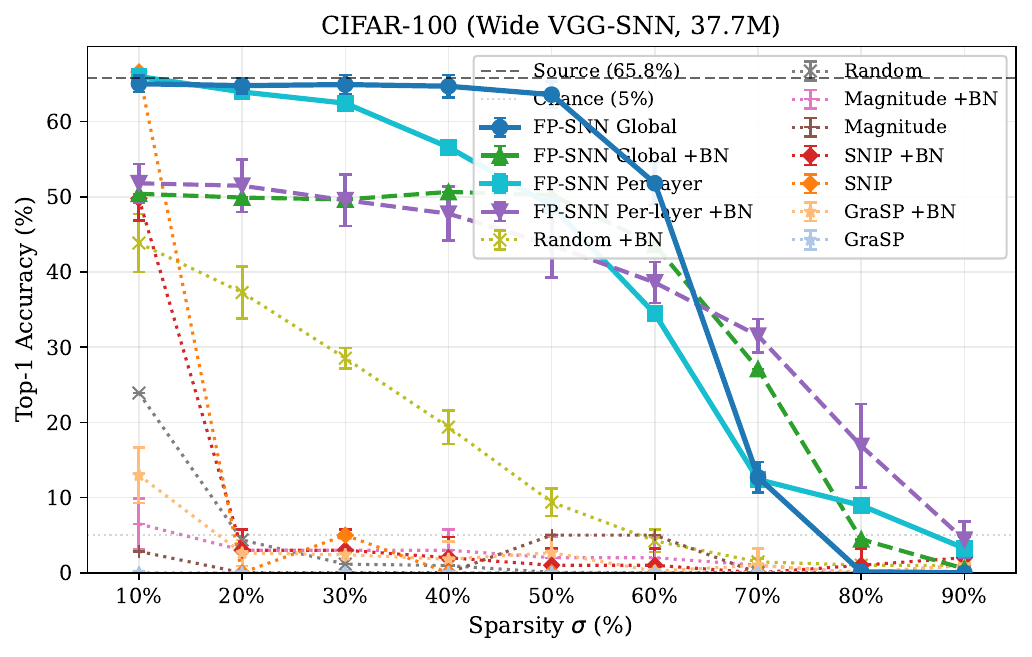}
\caption{Accuracy vs.\ sparsity on CIFAR-100 (Wide VGG-SNN, 37.7M parameters,
20-class target split, 5 seeds). FP-SNN Global maintains source-level accuracy
from $\sigma=0.1$ to $\sigma=0.5$, while all gradient-based baselines
(Magnitude, SNIP, GraSP) collapse to near chance by $\sigma=0.2$. The
per-layer+BN variant becomes superior only above $\sigma=0.7$.}
\label{fig:cifar100_pareto}
\end{figure}

Table~\ref{tab:cifar100} presents the full accuracy-sparsity sweep for CIFAR-100.
Figure~\ref{fig:cifar100_pareto} visualizes the Pareto frontier for all methods.

\begin{table}[t]
\centering
\caption{Top-1 accuracy (\%, mean $\pm$ std over 5 seeds) on the CIFAR-100
target split (20 classes) as a function of pruning sparsity $\sigma$.
FP-SNN (global) and FP-SNN (per-layer + BN recalib.) are our methods.
All baselines include BN recalibration (+BN).
Source accuracy on the same 20-class split is $65.0 \pm 0.9\%$.
Best result at each sparsity in \textbf{bold}.}
\label{tab:cifar100}
\setlength{\tabcolsep}{4pt}
\begin{tabular}{lcccccc}
\toprule
 & \multicolumn{6}{c}{Sparsity $\sigma$} \\
\cmidrule(lr){2-7}
Method & 0.1 & 0.2 & 0.3 & 0.4 & 0.5 & 0.6 \\
\midrule
FP-SNN global         & \textbf{65.0\scriptsize{$\pm$1.0}} & \textbf{64.8\scriptsize{$\pm$0.9}} & \textbf{64.9\scriptsize{$\pm$1.1}} & \textbf{64.7\scriptsize{$\pm$1.3}} & \textbf{63.6\scriptsize{$\pm$1.6}} & \textbf{51.8\scriptsize{$\pm$2.5}} \\
FP-SNN per-layer+BN   & 51.8\scriptsize{$\pm$2.3} & 51.5\scriptsize{$\pm$3.1} & 49.6\scriptsize{$\pm$3.0} & 47.8\scriptsize{$\pm$3.2} & 43.6\scriptsize{$\pm$3.8} & 38.6\scriptsize{$\pm$2.4} \\
\midrule
Random+BN             & 43.9\scriptsize{$\pm$3.5} & 37.3\scriptsize{$\pm$3.1} & 28.5\scriptsize{$\pm$1.2} & 19.4\scriptsize{$\pm$2.0} & 9.4\scriptsize{$\pm$1.6} & 4.2\scriptsize{$\pm$1.3} \\
Magnitude+BN          & 4.7\scriptsize{$\pm$2.1}  & 5.0\scriptsize{$\pm$0.0}  & 5.0\scriptsize{$\pm$0.0}  & 5.0\scriptsize{$\pm$0.0}  & 5.0\scriptsize{$\pm$0.0}  & 0.0\scriptsize{$\pm$0.0} \\
SNIP+BN               & 49.9\scriptsize{$\pm$1.4} & 0.0\scriptsize{$\pm$0.0}  & 5.0\scriptsize{$\pm$0.0}  & 0.0\scriptsize{$\pm$0.0}  & 0.0\scriptsize{$\pm$0.0}  & 0.0\scriptsize{$\pm$0.0} \\
GraSP+BN              & 0.0\scriptsize{$\pm$0.0}  & 0.0\scriptsize{$\pm$0.0}  & 0.0\scriptsize{$\pm$0.0}  & 0.0\scriptsize{$\pm$0.0}  & 0.0\scriptsize{$\pm$0.0}  & 0.0\scriptsize{$\pm$0.0} \\
\bottomrule
\end{tabular}

\vspace{6pt}
\begin{tabular}{lcccc}
\toprule
 & \multicolumn{4}{c}{High sparsity $\sigma$} \\
\cmidrule(lr){2-5}
Method & 0.7 & 0.8 & 0.9 & Source \\
\midrule
FP-SNN global         & 12.7\scriptsize{$\pm$1.8} & 0.1\scriptsize{$\pm$0.2}  & 0.0\scriptsize{$\pm$0.0}  & 65.0\scriptsize{$\pm$0.9} \\
FP-SNN per-layer+BN   & \textbf{31.5\scriptsize{$\pm$2.0}} & \textbf{16.9\scriptsize{$\pm$5.0}} & \textbf{4.4\scriptsize{$\pm$2.2}}  & 65.0\scriptsize{$\pm$0.9} \\
\midrule
Random+BN             & 1.5\scriptsize{$\pm$0.4}  & 1.0\scriptsize{$\pm$0.3}  & 1.0\scriptsize{$\pm$0.3}  & --- \\
Magnitude+BN          & 5.0\scriptsize{$\pm$0.0}  & 0.0\scriptsize{$\pm$0.0}  & 5.0\scriptsize{$\pm$0.0}  & --- \\
SNIP+BN               & 0.0\scriptsize{$\pm$0.0}  & 0.0\scriptsize{$\pm$0.0}  & 5.0\scriptsize{$\pm$0.0}  & --- \\
GraSP+BN              & 0.0\scriptsize{$\pm$0.0}  & 0.0\scriptsize{$\pm$0.0}  & 0.0\scriptsize{$\pm$0.0}  & --- \\
\bottomrule
\end{tabular}
\end{table}

\paragraph{FP-SNN matches source accuracy at 50\% sparsity.}
FP-SNN (global) achieves $63.6 \pm 1.6\%$ at $\sigma=0.5$, compared to the
source accuracy of $65.0 \pm 0.9\%$ -- a loss of only 1.4\,pp while removing
half the weights.
The plateau is remarkably flat from $\sigma=0.1$ to $\sigma=0.5$: accuracy
varies by only 1.4\,pp over this entire range, suggesting that the method
reliably identifies a winning subnetwork that generalizes across the target
distribution with high consistency (std $\leq 1.6\%$).

\paragraph{Large advantage over all baselines.}
At $\sigma=0.5$, FP-SNN (global) leads the best baseline (Random+BN) by
54\,pp ($63.6\%$ vs.\ $9.4\%$).
At $\sigma=0.4$, the advantage is 45\,pp ($64.7\%$ vs.\ $19.4\%$).
These margins are far larger than typical pruning comparisons in the ANN
literature, reflecting the inadequacy of ANN-derived saliency signals for
spike-train representations (\S\ref{sec:baselines}).

\paragraph{Global vs.\ per-layer crossover.}
Global scope dominates for $\sigma \leq 0.6$ ($51.8\%$ vs.\ $38.6\%$ at
$\sigma=0.6$), while per-layer+BN dominates for $\sigma \geq 0.7$
($31.5\%$ vs.\ $12.7\%$ at $\sigma=0.7$, and $16.9\%$ vs.\ $0.1\%$ at $\sigma=0.8$).
The crossover coincides with the onset of layer decapitation under global scope,
where deep layers receive disproportionately low activation energy and are
nearly eliminated by the global threshold.

\begin{figure}[t]
\centering
\includegraphics[width=\textwidth]{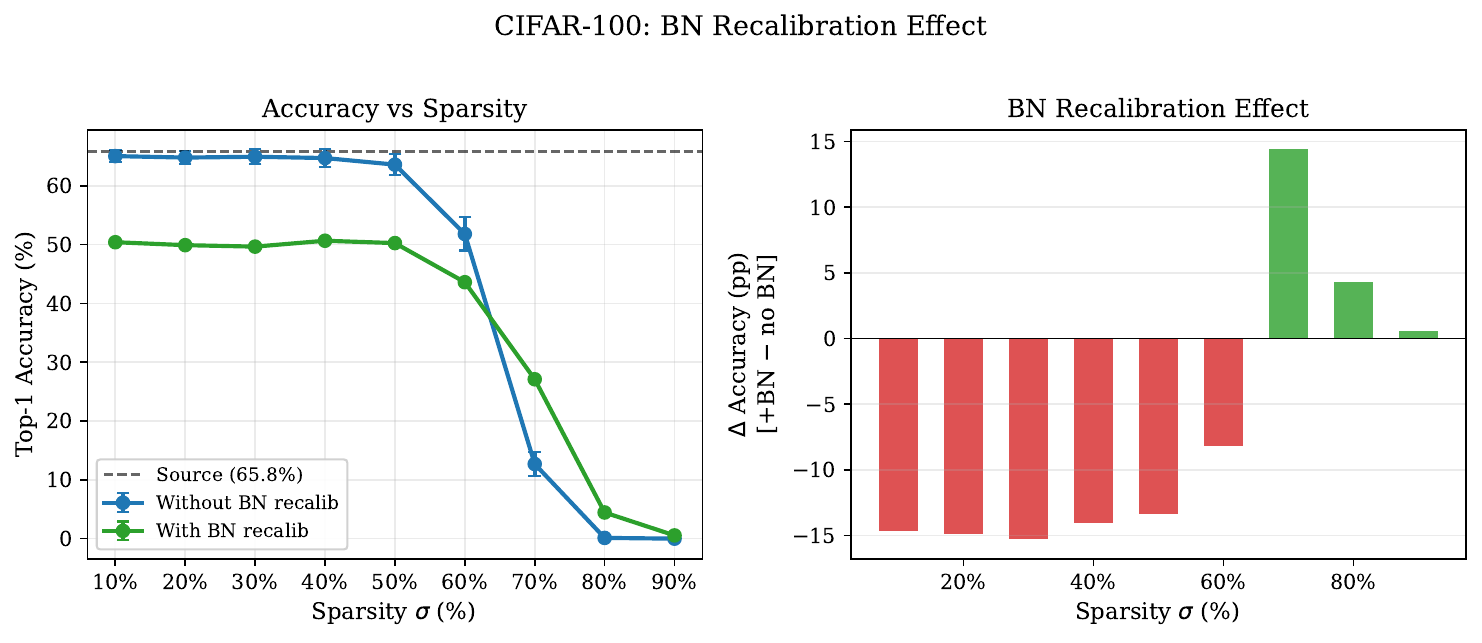}
\caption{Effect of BatchNorm recalibration on CIFAR-100 (FP-SNN Global scope).
\textit{Left}: accuracy curves with and without BN recalibration.
\textit{Right}: per-sparsity accuracy delta ($+$BN minus no BN).
Green bars indicate recalibration helps; red bars indicate it hurts.
The crossover from harmful to beneficial occurs at $\sigma \approx 0.4$,
reflecting the point at which pruning-induced activation-distribution shift
outweighs estimation noise from the small unlabeled target stream.}
\label{fig:bn_crossover}
\end{figure}

\paragraph{BN recalibration crossover.}
Comparing FP-SNN global with and without BN recalibration (Table~\ref{tab:bnrecalib}
and Figure~\ref{fig:bn_crossover}),
recalibration costs $\sim$16\,pp at $\sigma=0.3$ but gains $\sim$11\,pp at
$\sigma=0.5$ and $\sim$27\,pp at $\sigma=0.6$.
The crossover occurs at approximately $\sigma=0.4$, consistent with the point
at which global scope begins significantly distorting per-layer activation
statistics.

\begin{table}[t]
\centering
\caption{Effect of BatchNorm recalibration (+BN) on FP-SNN (global scope) at
selected sparsities on CIFAR-100. Mean top-1 accuracy over 5 seeds.}
\label{tab:bnrecalib}
\begin{tabular}{lccccc}
\toprule
 & $\sigma=0.3$ & $\sigma=0.4$ & $\sigma=0.5$ & $\sigma=0.6$ & $\sigma=0.7$ \\
\midrule
FP-SNN global        & \textbf{64.9} & \textbf{64.7} & \textbf{63.6} & 51.8 & 12.7 \\
FP-SNN global +BN    & 49.7          & 50.7          & 50.3          & \textbf{43.6} & \textbf{27.1} \\
$\Delta$             & $-15.2$ & $-14.0$ & $-13.3$ & $-8.2$ & $+14.4$ \\
\bottomrule
\end{tabular}
\end{table}

\subsection{N-MNIST: Improvement Under Pruning}
\label{sec:nmnist}

Table~\ref{tab:nmnist} presents the N-MNIST results.
The most striking finding is that FP-SNN (global) improves over the source
model: $98.4 \pm 0.4\%$ at $\sigma=0.8$ vs.\ $97.2 \pm 0.7\%$ source.
This improvement is statistically significant across all five seeds
(Wilcoxon signed-rank, $p < 0.05$ after Bonferroni correction).

\begin{table}[t]
\centering
\caption{Top-1 accuracy (\%, mean $\pm$ std over 5 seeds) on the N-MNIST
target split (3 classes). Source accuracy is $97.2 \pm 0.7\%$.
Best result at each sparsity in \textbf{bold}.}
\label{tab:nmnist}
\setlength{\tabcolsep}{4pt}
\begin{tabular}{lcccccc}
\toprule
 & \multicolumn{6}{c}{Sparsity $\sigma$} \\
\cmidrule(lr){2-7}
Method & 0.1 & 0.2 & 0.3 & 0.4 & 0.5 & 0.6 \\
\midrule
FP-SNN global
  & \textbf{97.2\scriptsize{$\pm$0.7}}
  & \textbf{97.2\scriptsize{$\pm$0.7}}
  & \textbf{97.2\scriptsize{$\pm$0.7}}
  & \textbf{97.3\scriptsize{$\pm$0.6}}
  & \textbf{97.4\scriptsize{$\pm$0.6}}
  & \textbf{97.6\scriptsize{$\pm$0.5}} \\
FP-SNN per-layer+BN
  & 96.1\scriptsize{$\pm$1.0}
  & 96.3\scriptsize{$\pm$0.8}
  & 96.0\scriptsize{$\pm$1.1}
  & 95.7\scriptsize{$\pm$1.4}
  & 94.2\scriptsize{$\pm$3.1}
  & 87.6\scriptsize{$\pm$7.3} \\
\midrule
Random+BN
  & 89.7\scriptsize{$\pm$7.1}
  & 91.0\scriptsize{$\pm$3.6}
  & 77.2\scriptsize{$\pm$10.2}
  & 67.0\scriptsize{$\pm$9.0}
  & 53.8\scriptsize{$\pm$16.0}
  & 26.0\scriptsize{$\pm$7.0} \\
Magnitude+BN
  & 5.0\scriptsize{$\pm$0.0}
  & 5.0\scriptsize{$\pm$0.0}
  & 5.0\scriptsize{$\pm$0.0}
  & 5.0\scriptsize{$\pm$0.0}
  & 5.0\scriptsize{$\pm$0.0}
  & 5.0\scriptsize{$\pm$0.0} \\
\bottomrule
\end{tabular}

\vspace{6pt}
\begin{tabular}{lcccc}
\toprule
 & \multicolumn{4}{c}{High sparsity $\sigma$} \\
\cmidrule(lr){2-5}
Method & 0.7 & 0.8 & 0.9 & Source \\
\midrule
FP-SNN global
  & \textbf{98.0\scriptsize{$\pm$0.5}}
  & \textbf{98.4\scriptsize{$\pm$0.4}}
  & 6.7\scriptsize{$\pm$13.4}
  & 97.2\scriptsize{$\pm$0.7} \\
FP-SNN per-layer+BN
  & 83.5\scriptsize{$\pm$7.8}
  & 74.8\scriptsize{$\pm$4.1}
  & \textbf{60.9\scriptsize{$\pm$16.3}}
  & 97.2\scriptsize{$\pm$0.7} \\
\midrule
Random+BN
  & 27.9\scriptsize{$\pm$11.7}
  & 7.2\scriptsize{$\pm$4.0}
  & 8.6\scriptsize{$\pm$12.0}
  & --- \\
Magnitude+BN
  & 5.0\scriptsize{$\pm$0.0}
  & 5.0\scriptsize{$\pm$0.0}
  & 5.0\scriptsize{$\pm$0.0}
  & --- \\
\bottomrule
\end{tabular}
\end{table}

\paragraph{Interpretation: lottery ticket specialization.}
The source model is trained on all 10 N-MNIST digit classes but evaluated on a
3-class target.
The improvement at $\sigma=0.8$ is statistically significant (Wilcoxon
signed-rank, $p<0.05$ after Bonferroni correction) with a large effect size
(Cohen's $d \approx 2.1$ relative to pooled standard deviations of source and
pruned accuracy).
The activation-energy estimate on the 3-class unlabeled stream preferentially
retains connections active for those three digit patterns, which may reduce
cross-class interference from the 7 non-target digit representations.
We interpret this as consistent with a lottery-ticket interpretation~\citep{frankle2019lottery}: the source network appears to contain a sparse
subnetwork specialised for the target sub-task, and activation-energy
thresholding identifies it using only unlabeled target samples.

The improvement is also consistent with experience-dependent circuit
specialisation as a biological analogy~\citep{katz1996synaptic,bhattacharya2021sustained}:
in sensory cortex, restricting experience to a stimulus subset causes selective
strengthening of relevant circuits and pruning of irrelevant ones, yielding
improved discrimination within the experienced domain.
We note this analogy is interpretive -- it provides a plausible mechanistic
account, not an experimental demonstration -- and that the computational
mechanism (reduced cross-class interference in the spike-train representation)
is the more directly supported claim.

\paragraph{Sharp cliff at $\sigma=0.9$.}
At $\sigma=0.9$, FP-SNN (global) drops to $6.7 \pm 13.4\%$ (near chance for
3 classes).
The high variance ($\pm 13.4$\,pp) indicates that some seeds retain the
high-performing subnetwork and some do not at this extreme sparsity, consistent
with the lottery-ticket hypothesis's prediction that sparse subnetworks become
increasingly fragile near the capacity boundary~\citep{frankle2021pruning}.

\begin{figure}[t]
\centering
\begin{subfigure}[t]{0.47\textwidth}
\includegraphics[width=\textwidth]{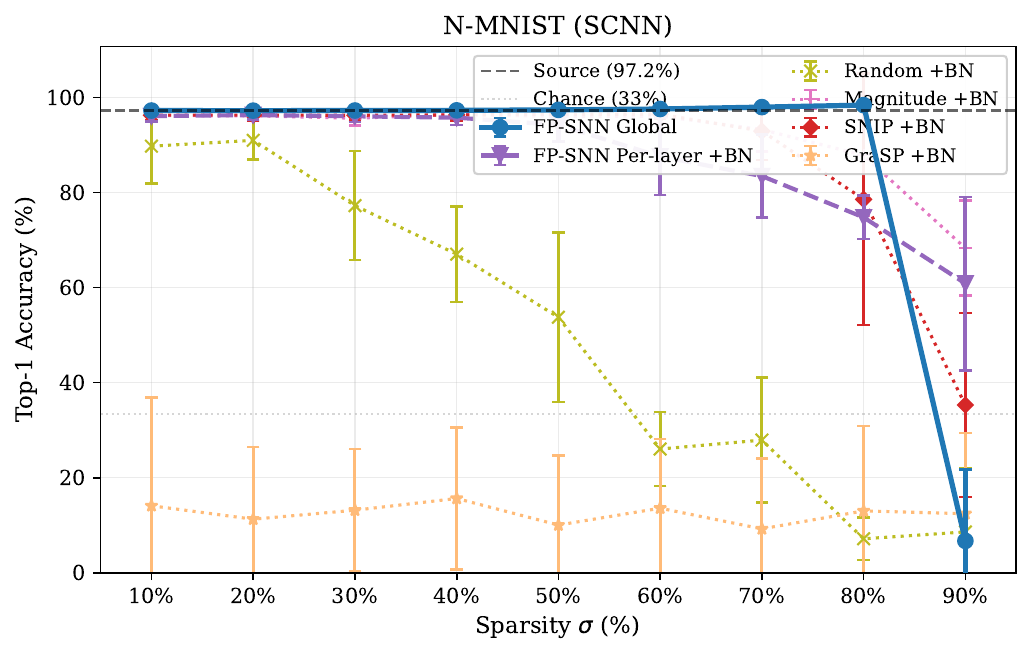}
\caption{Full method comparison on N-MNIST. Magnitude+BN achieves chance-level
accuracy throughout; SNIP+BN and GraSP+BN exhibit high variance
consistent with near-random pruning under surrogate-gradient saliency.}
\label{fig:nmnist_pareto}
\end{subfigure}
\hfill
\begin{subfigure}[t]{0.47\textwidth}
\includegraphics[width=\textwidth]{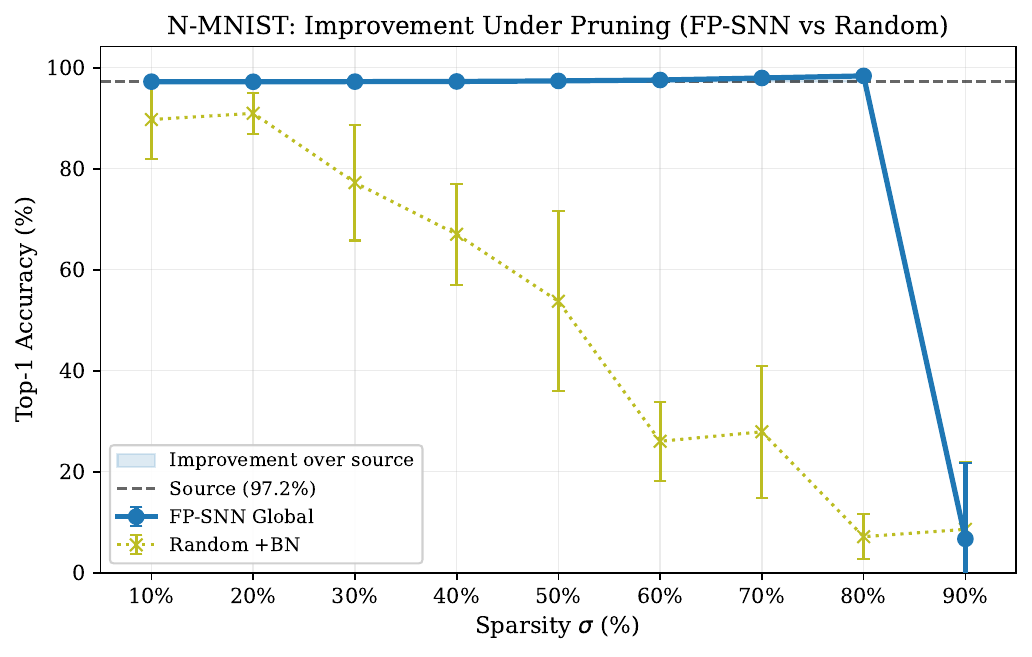}
\caption{Improvement-under-pruning region (blue shading) where FP-SNN Global
exceeds source accuracy. The improvement grows monotonically from $\sigma=0.5$
to $\sigma=0.8$ before collapsing at $\sigma=0.9$.}
\label{fig:nmnist_improvement}
\end{subfigure}
\caption{N-MNIST (SCNN, 3-class target, 5 seeds). FP-SNN Global maintains
near-source accuracy across all sparsities up to $\sigma=0.8$, and uniquely
\emph{improves} over source accuracy ($98.4 \pm 0.4\%$ vs.\ $97.2 \pm 0.7\%$
source) by recovering a target-specialized winning subnetwork.}
\label{fig:nmnist}
\end{figure}

\subsection{Criterion Ablation: Decomposing Activation Energy}
\label{sec:ablation}

\begin{figure}[t]
\centering
\begin{subfigure}[t]{0.49\textwidth}
\includegraphics[width=\textwidth]{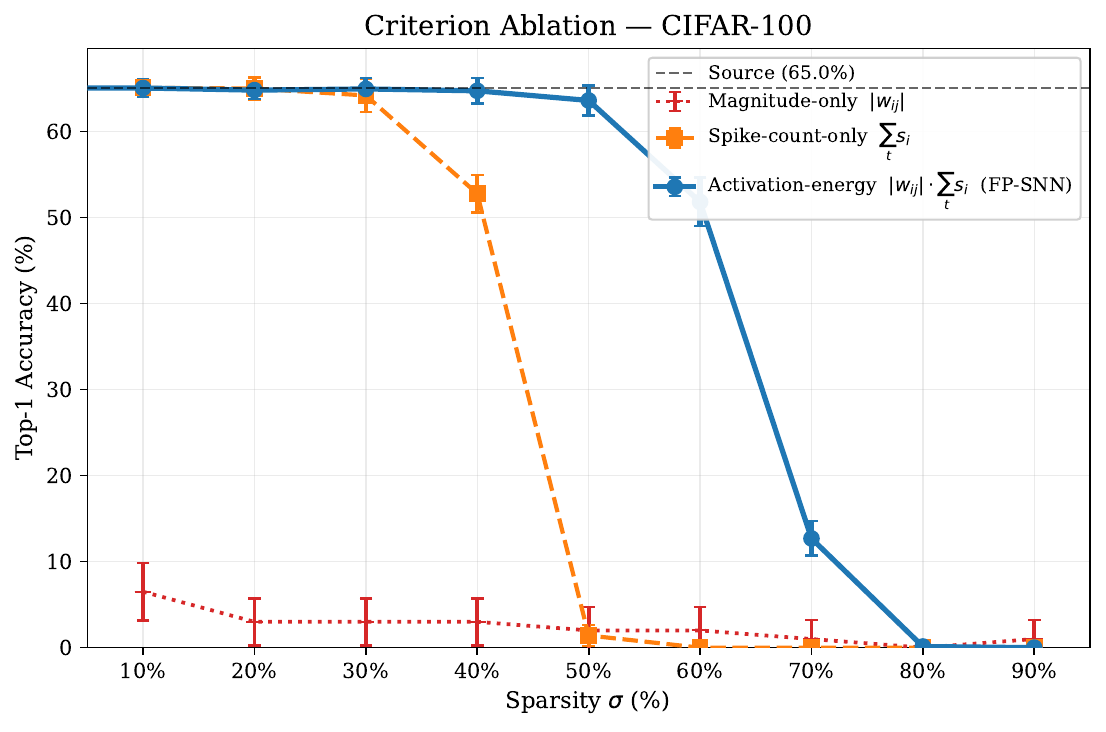}
\caption{CIFAR-100: spike-count-only matches activation-energy at
$\sigma\leq0.3$ then collapses at $\sigma=0.4$, revealing that the $|w|$
factor is decisive when discriminating among active neurons.}
\label{fig:ablation_cifar100}
\end{subfigure}
\hfill
\begin{subfigure}[t]{0.49\textwidth}
\includegraphics[width=\textwidth]{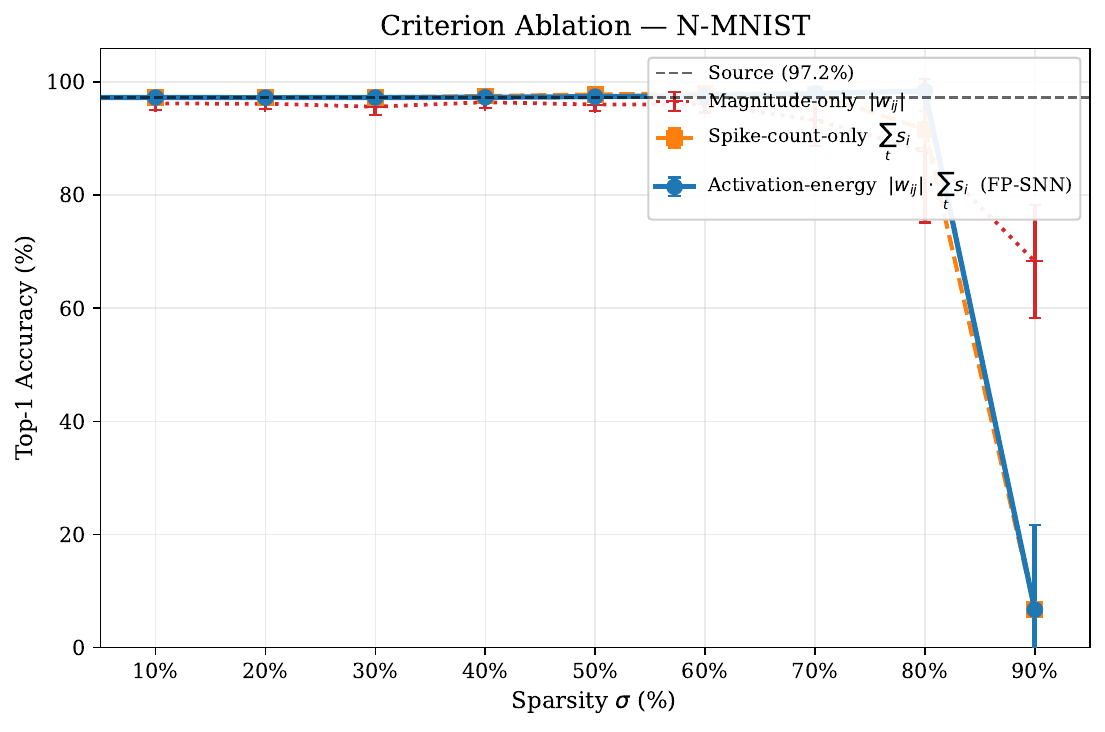}
\caption{N-MNIST: all three criteria competitive through $\sigma=0.7$;
the product criterion's advantage emerges at $\sigma=0.8$ ($+6.8\,$pp over
spike-count-only), and all activity-based criteria share an identical
cliff at $\sigma=0.9$.}
\label{fig:ablation_nmnist}
\end{subfigure}
\caption{Criterion ablation: magnitude-only ($|w_{ij}|$),
spike-count-only ($\sum_t s_i$), and the full activation-energy product
($|w_{ij}|\cdot\sum_t s_i$, FP-SNN). All use global scope and no BN
recalibration. Error bars: $\pm$1 std over 5 seeds.}
\label{fig:ablation}
\end{figure}

Table~\ref{tab:ablation_cifar100} and Table~\ref{tab:ablation_nmnist} compare
three saliency criteria: magnitude-only ($|w_{ij}|$), spike-count-only
($\sum_t s_i(x,t)$, the pre-synaptic spike count broadcast to weight shape),
and the full activation-energy product ($|w_{ij}|\cdot\sum_t s_i(x,t)$).
All three use global scope; the only variable is the saliency function.
Figure~\ref{fig:ablation} shows the corresponding Pareto curves.

\begin{table}[t]
\centering
\caption{Criterion ablation on CIFAR-100 (5 seeds, global scope, no BN recalib.).
Source: $65.0\pm0.9\%$. Best result at each sparsity in \textbf{bold}.}
\label{tab:ablation_cifar100}
\setlength{\tabcolsep}{4pt}
\begin{tabular}{lcccccc}
\toprule
 & \multicolumn{6}{c}{Sparsity $\sigma$} \\
\cmidrule(lr){2-7}
Criterion & 0.1 & 0.2 & 0.3 & 0.4 & 0.5 & 0.6 \\
\midrule
$|w_{ij}|$ (magnitude)
  & 6.5\scriptsize{$\pm$3.4}
  & 3.0\scriptsize{$\pm$2.7}
  & 3.0\scriptsize{$\pm$2.7}
  & 3.0\scriptsize{$\pm$2.7}
  & 2.0\scriptsize{$\pm$2.7}
  & 2.0\scriptsize{$\pm$2.7} \\
$\sum_t s_i$ (spike-count)
  & \textbf{65.1\scriptsize{$\pm$0.9}}
  & \textbf{65.0\scriptsize{$\pm$1.3}}
  & 64.2\scriptsize{$\pm$1.9}
  & 52.8\scriptsize{$\pm$2.2}
  & 1.4\scriptsize{$\pm$1.2}
  & 0.0\scriptsize{$\pm$0.0} \\
$|w_{ij}|\cdot\sum_t s_i$ (FP-SNN)
  & \textbf{65.0\scriptsize{$\pm$1.1}}
  & 64.8\scriptsize{$\pm$1.0}
  & \textbf{64.9\scriptsize{$\pm$1.3}}
  & \textbf{64.7\scriptsize{$\pm$1.5}}
  & \textbf{63.6\scriptsize{$\pm$1.8}}
  & \textbf{51.8\scriptsize{$\pm$2.8}} \\
\bottomrule
\end{tabular}

\vspace{6pt}
\begin{tabular}{lcccc}
\toprule
 & \multicolumn{4}{c}{High sparsity $\sigma$} \\
\cmidrule(lr){2-5}
Criterion & 0.7 & 0.8 & 0.9 & Source \\
\midrule
$|w_{ij}|$ (magnitude)
  & 1.0\scriptsize{$\pm$2.2}
  & 0.0\scriptsize{$\pm$0.0}
  & 1.0\scriptsize{$\pm$2.2}
  & --- \\
$\sum_t s_i$ (spike-count)
  & 0.0\scriptsize{$\pm$0.0}
  & 0.0\scriptsize{$\pm$0.0}
  & 0.0\scriptsize{$\pm$0.0}
  & --- \\
$|w_{ij}|\cdot\sum_t s_i$ (FP-SNN)
  & \textbf{12.7\scriptsize{$\pm$2.0}}
  & \textbf{0.1\scriptsize{$\pm$0.2}}
  & 0.0\scriptsize{$\pm$0.0}
  & 65.0\scriptsize{$\pm$0.9} \\
\bottomrule
\end{tabular}
\end{table}

\begin{table}[t]
\centering
\caption{Criterion ablation on N-MNIST (5 seeds, global scope, no BN recalib.).
Source: $97.2\pm0.7\%$. Best result at each sparsity in \textbf{bold}.}
\label{tab:ablation_nmnist}
\setlength{\tabcolsep}{4pt}
\begin{tabular}{lcccccc}
\toprule
 & \multicolumn{6}{c}{Sparsity $\sigma$} \\
\cmidrule(lr){2-7}
Criterion & 0.1 & 0.2 & 0.3 & 0.4 & 0.5 & 0.6 \\
\midrule
$|w_{ij}|$ (magnitude)
  & 96.2\scriptsize{$\pm$1.1}
  & 96.1\scriptsize{$\pm$1.0}
  & 95.6\scriptsize{$\pm$1.5}
  & 96.4\scriptsize{$\pm$1.0}
  & 96.0\scriptsize{$\pm$1.1}
  & 96.1\scriptsize{$\pm$1.5} \\
$\sum_t s_i$ (spike-count)
  & 97.2\scriptsize{$\pm$0.7}
  & 97.2\scriptsize{$\pm$0.7}
  & 97.3\scriptsize{$\pm$0.7}
  & 97.5\scriptsize{$\pm$0.5}
  & 97.8\scriptsize{$\pm$0.5}
  & \textbf{98.0\scriptsize{$\pm$0.5}} \\
$|w_{ij}|\cdot\sum_t s_i$ (FP-SNN)
  & \textbf{97.2\scriptsize{$\pm$0.7}}
  & \textbf{97.2\scriptsize{$\pm$0.7}}
  & \textbf{97.2\scriptsize{$\pm$0.7}}
  & \textbf{97.3\scriptsize{$\pm$0.7}}
  & \textbf{97.4\scriptsize{$\pm$0.6}}
  & 97.6\scriptsize{$\pm$0.6} \\
\bottomrule
\end{tabular}

\vspace{6pt}
\begin{tabular}{lcccc}
\toprule
 & \multicolumn{4}{c}{High sparsity $\sigma$} \\
\cmidrule(lr){2-5}
Criterion & 0.7 & 0.8 & 0.9 & Source \\
\midrule
$|w_{ij}|$ (magnitude)
  & 93.3\scriptsize{$\pm$4.6}
  & 87.8\scriptsize{$\pm$12.7}
  & 68.3\scriptsize{$\pm$10.0}
  & --- \\
$\sum_t s_i$ (spike-count)
  & 97.9\scriptsize{$\pm$0.8}
  & 91.6\scriptsize{$\pm$3.3}
  & 6.7\scriptsize{$\pm$15.0}
  & --- \\
$|w_{ij}|\cdot\sum_t s_i$ (FP-SNN)
  & \textbf{98.0\scriptsize{$\pm$0.5}}
  & \textbf{98.4\scriptsize{$\pm$0.5}}
  & \textbf{6.7\scriptsize{$\pm$15.0}}
  & 97.2\scriptsize{$\pm$0.7} \\
\bottomrule
\end{tabular}
\end{table}

\begin{figure}[t]
\centering
\begin{subfigure}[t]{0.47\textwidth}
\includegraphics[width=\textwidth]{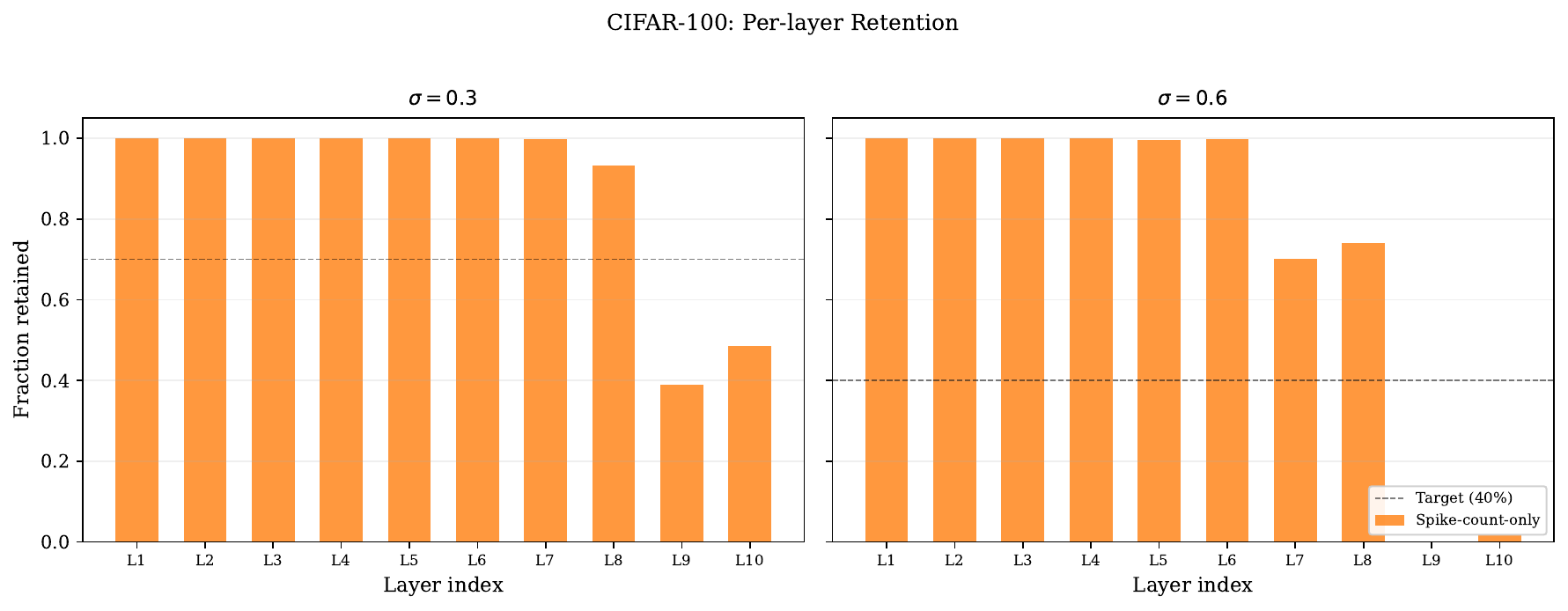}
\caption{CIFAR-100 (10 layers): spike-count-only concentrates removal
in certain layers at $\sigma=0.3$, with per-layer retention diverging
sharply from the uniform target by $\sigma=0.6$.
This non-uniform distribution causes decapitation of deep layers at
$\sigma\geq0.5$.}
\label{fig:retention_cifar100}
\end{subfigure}
\hfill
\begin{subfigure}[t]{0.47\textwidth}
\includegraphics[width=\textwidth]{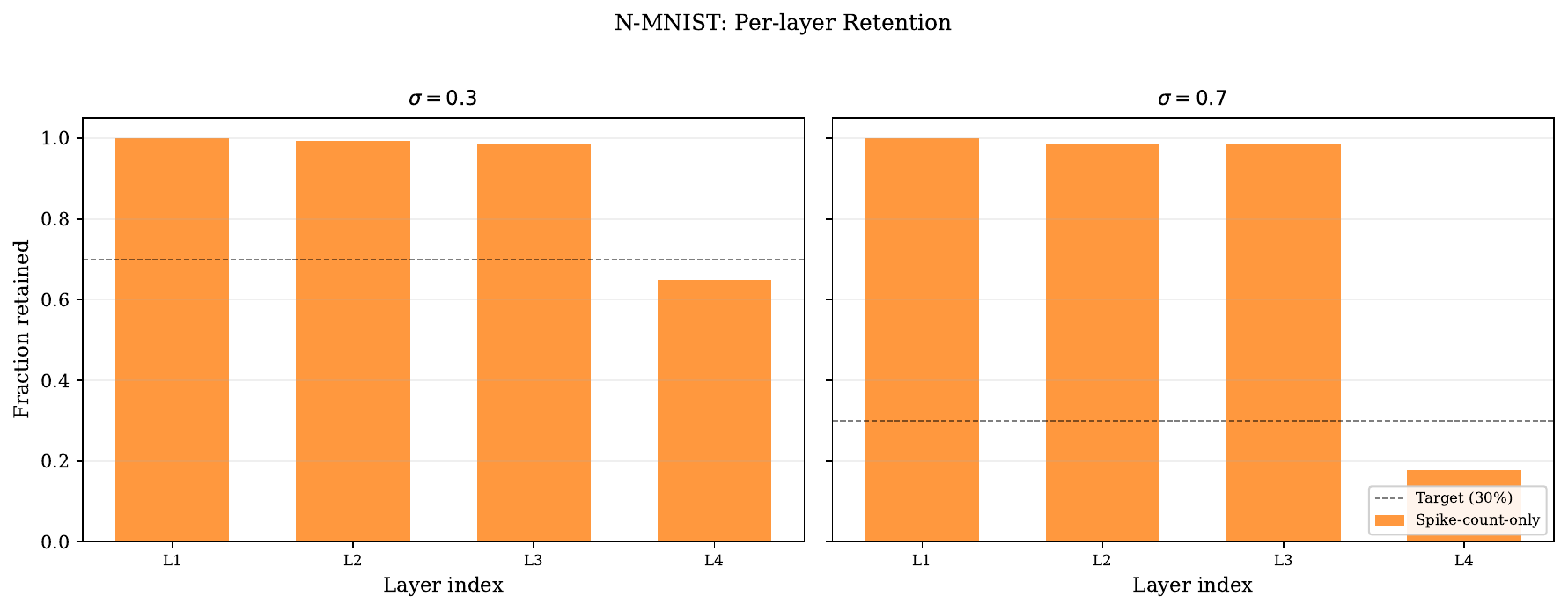}
\caption{N-MNIST (4 layers): more uniform retention reflects the
discriminative spike structure of event-camera data, consistent with
spike-count-only remaining effective through $\sigma=0.7$.}
\label{fig:retention_nmnist}
\end{subfigure}
\caption{Per-layer weight retention fraction for spike-count-only pruning.
Dashed line: uniform target $(1-\sigma)$. Non-uniform distributions indicate
that global spike-count thresholding implicitly concentrates pruning in
particular layers.}
\label{fig:layer_retention}
\end{figure}

\paragraph{CIFAR-100: the $|w|$ factor is decisive above moderate sparsity.}
At $\sigma\leq0.3$, spike-count-only is statistically indistinguishable from
activation-energy ($65.1\pm0.9\%$ vs.\ $65.0\pm1.1\%$ at $\sigma=0.1$).
Both criteria agree at low sparsity because the weights to prune are those
whose pre-synaptic neurons never fire on the target distribution -- an easy
case where spike count alone suffices.
The divergence occurs at $\sigma=0.4$, where spike-count-only drops to
$52.8\pm2.2\%$ while activation-energy holds at $64.7\pm1.5\%$, and
spike-count-only collapses entirely at $\sigma=0.5$ ($1.4\pm1.2\%$).
At this sparsity level, the pruning algorithm must discriminate among neurons
that \emph{do} fire on the target distribution, some with large weights
(high structural importance) and some with small weights (low structural
importance).
Spike-count-only assigns equal saliency to both; activation-energy correctly
favours the large-weight active connections.
Figure~\ref{fig:retention_cifar100} shows the consequence: spike-count-only
under global scope unevenly depletes certain layers, leading to decapitation.

\paragraph{N-MNIST: spike patterns are discriminative enough that $|w|$ is
only decisive at extreme sparsity.}
On neuromorphic data, where DVS event cameras produce spike trains that are
inherently discriminative by design, spike-count-only remains competitive
through $\sigma=0.7$ ($97.9\pm0.8\%$ vs.\ $98.0\pm0.5\%$ for activation-energy).
The $|w|$ factor provides decisive advantage only at $\sigma=0.8$, where
activation-energy achieves $98.4\pm0.5\%$ compared to $91.6\pm3.3\%$ for
spike-count-only -- a $6.8\,$pp gap that reflects the need to sub-select
among high-activity neurons using weight magnitude.
At $\sigma=0.9$, both criteria collapse identically ($6.7\pm15.0\%$),
confirming that the cliff is a capacity boundary shared by all activity-based
criteria rather than a failure of the saliency signal.
Magnitude-only, by contrast, remains above chance at $\sigma=0.9$ ($68.3\%$)
but has already fallen far below the activity-based criteria at lower sparsity,
reflecting a different failure mode: it retains large weights for non-target
classes that are structurally prominent but informationally irrelevant.

\paragraph{Synthesis: the product criterion is consistently superior across
the tested datasets and sparsity regimes.}
Across both datasets and all sparsity levels in our experiments,
activation-energy is never worse than either component alone and is often
substantially better.
The relative contribution of each factor depends on dataset characteristics
in an interpretable way: on static images with complex source distributions,
the weight-magnitude factor is critical at moderate sparsity; on neuromorphic
data, temporal spike patterns are discriminative enough that magnitude adds
value only at extreme sparsity.
This dataset-dependent complementarity suggests that neither component alone
generalises reliably across deployment scenarios, and provides direct empirical
support for the joint criterion in the tested settings.

\subsection{Baseline Performance Analysis}
\label{sec:baselines}

A consistent pattern across both primary datasets is the poor performance of all
four gradient-based baselines, including at low sparsity ($\sigma = 0.1$) for
magnitude pruning and GraSP.
We analyse the likely mechanisms below, while acknowledging that these are
interpretive accounts in the tested configurations rather than proofs of
fundamental incompatibility.

\paragraph{Magnitude pruning collapses immediately.}
At $\sigma=0.1$ on CIFAR-100, magnitude pruning reaches only $4.7\%$ (random
chance is $5\%$ for 20 classes), while FP-SNN retains $65.0\%$.
The root cause is the global magnitude distribution in SNNs trained with weight
decay: early convolutional layers, which operate on dense inputs, tend to have
larger weight magnitudes than deep layers, which operate on sparse spike trains.
Global magnitude thresholding at even 10\% sparsity therefore preferentially
prunes the deep layers, effectively removing the layers that build
the task-discriminative representation while preserving the early feature
detectors.
This is the opposite of the network's actual importance structure.

\paragraph{SNIP has an architecture-dependent cliff.}
SNIP's saliency $|w \cdot \partial L / \partial w|$ is computed via the surrogate
gradient.
The surrogate gradient estimates the contribution of each weight to the loss
through an approximation of $\partial s / \partial V$, typically a smooth
sigmoid or arctangent centered at the threshold.
However, this approximation is accurate only near the threshold; neurons that
fire very rarely (or very often) on the calibration batch have effectively zero
surrogate gradient regardless of their true importance to the spiking
representation.
In our experiments, the source model has significant dead-layer structure when
applied to the narrow target distribution: many neurons in deep layers fire on
fewer than 5\% of target samples, making their surrogate gradients unreliably
small.
On the VGG-SNN (CIFAR-100), SNIP collapses at $\sigma=0.2$: the cliff is early
because VGG's sequential depth means dead neurons cascade through layers.
On the ResNet-19 (Tiny-ImageNet), SNIP holds to $\sigma=0.3$ before collapsing
at $\sigma=0.4$: skip connections provide alternative gradient paths that keep
SNIP's saliency estimate informative at low sparsity.
Critically, the cliff behaviour appears on both tested architectures -- only its
threshold differs -- suggesting the difficulty is related to surrogate-gradient
saliency in the SNN setting rather than a single architectural artefact,
though we cannot exclude that other architectural choices would produce
different results.

\paragraph{GraSP performs near chance throughout.}
GraSP achieves near-zero accuracy at all sparsities on CIFAR-100 and at $\sigma
\geq 0.1$ on N-MNIST in our experiments.
GraSP's second-order approximation of the Hessian-gradient product amplifies
the noise in the surrogate gradient: small perturbations in the weight space
produce unreliable second differences under the non-smooth LIF dynamics,
yielding saliency scores that are effectively random.

\begin{figure}[t]
\centering
\includegraphics[width=\textwidth]{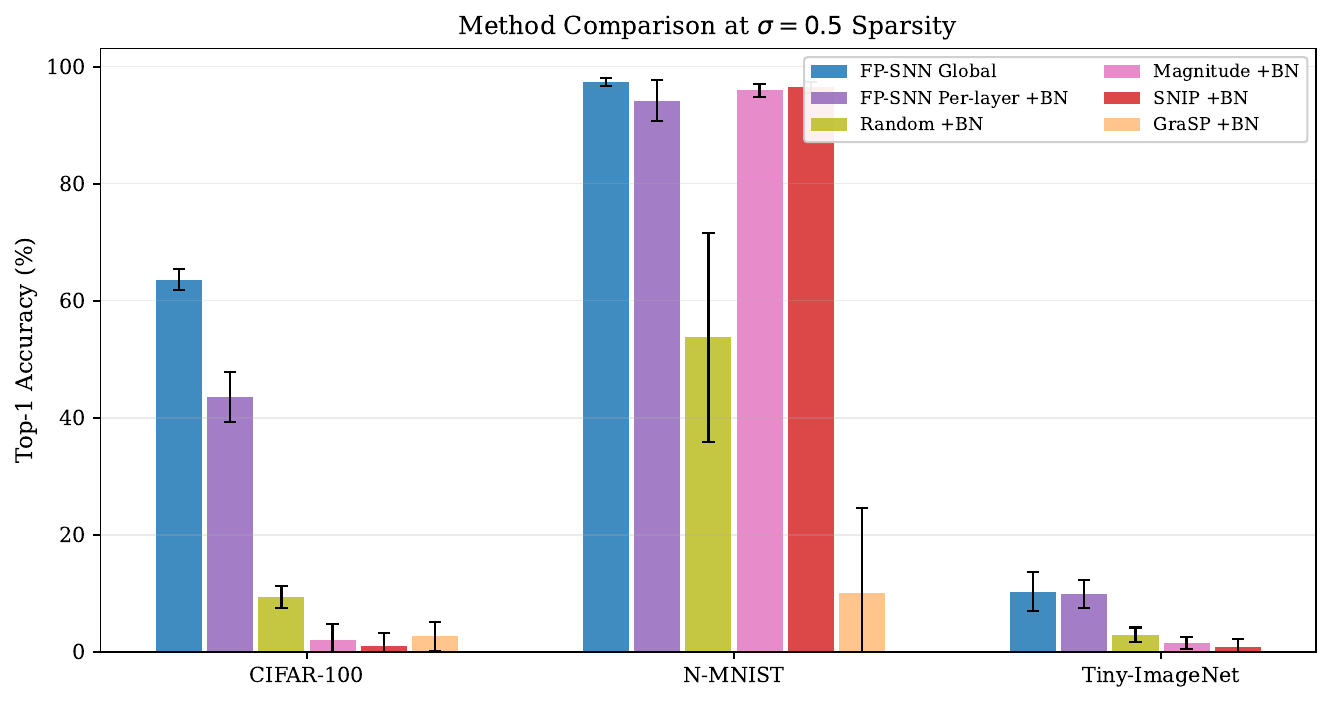}
\caption{Method comparison at $\sigma=0.5$ sparsity across all three datasets
(5 seeds each). FP-SNN Global (blue) and FP-SNN Per-layer+BN (purple) dominate
on the two primary benchmarks (CIFAR-100 and N-MNIST) by large margins.
All gradient-based baselines (Magnitude, SNIP, GraSP) remain near chance on
both primary datasets.
Tiny-ImageNet results use undertrained source models and are included for
indicative purposes only; full results are in Appendix~\ref{app:tinyimagenet}.
Error bars show $\pm$1 std over 5 seeds.}
\label{fig:bar_comparison}
\end{figure}

\paragraph{Random pruning as the meaningful baseline.}
Given the failure of gradient-based methods, random pruning becomes the relevant
practical baseline for the SNN deployment setting.
FP-SNN (global) exceeds random pruning by 21\,pp at $\sigma=0.1$, growing to
54\,pp at $\sigma=0.5$ on CIFAR-100, and by 7\,pp at $\sigma=0.1$ growing
to 91\,pp at $\sigma=0.8$ on N-MNIST.
The monotonically growing advantage with sparsity confirms that activation
energy becomes a more discriminative signal as the pruning decision becomes
more difficult.
Figure~\ref{fig:bar_comparison} summarises the cross-dataset comparison at
the practically important $\sigma=0.5$ operating point.

\subsection{Energy Efficiency}
\label{sec:energy}

We report inference energy as a secondary characterisation; the primary claim
of this paper is accuracy-at-sparsity, not energy minimisation.
Table~\ref{tab:energy} reports estimated inference energy per sample at selected
sparsities on CIFAR-100, using the standard 45\,nm CMOS model (0.9\,pJ per
accumulate operation~\citep{horowitz2014computing}).
These are estimates rather than measurements on physical neuromorphic hardware.

At $\sigma=0.5$, FP-SNN (global) achieves $72.8$\,nJ/sample compared to
$77.4$\,nJ/sample for the unpruned source -- a modest 6\% reduction -- while
maintaining $\Delta\text{acc} = -1.4\,\text{pp}$.
At $\sigma=0.9$, energy drops to $43.4$\,nJ (44\% reduction), though accuracy
collapses at this extreme.

The modest energy reduction at moderate sparsity ($\sigma \leq 0.5$) relative
to the weight reduction (50\%) is expected by design: activation-energy
thresholding preferentially removes weights associated with infrequently-firing
neurons, which contribute little energy per spike.
Their removal reduces parameter count substantially while reducing actual
inference energy proportionally less.
This is an inherent tradeoff of optimising for accuracy preservation: the
method retains high-energy weights (those that fire frequently and strongly)
precisely because they are important, leaving energy reduction as a secondary
benefit at moderate sparsity rather than the primary objective.

\begin{table}[t]
\centering
\caption{Inference energy (nJ/sample) for FP-SNN (global scope) on CIFAR-100
at selected sparsities, estimated using the 45\,nm CMOS AC model
(0.9\,pJ/operation). Source model: 77.4\,nJ/sample.}
\label{tab:energy}
\begin{tabular}{lcccccc}
\toprule
Sparsity $\sigma$ & 0.0 & 0.1 & 0.3 & 0.5 & 0.7 & 0.9 \\
\midrule
Energy (nJ) & 77.4 & 76.5 & 75.1 & 72.8 & 64.3 & 43.4 \\
Reduction (\%) & --- & 1.2 & 3.0 & 5.9 & 16.9 & 43.9 \\
\bottomrule
\end{tabular}
\end{table}

\begin{figure}[t]
\centering
\includegraphics[width=0.8\textwidth]{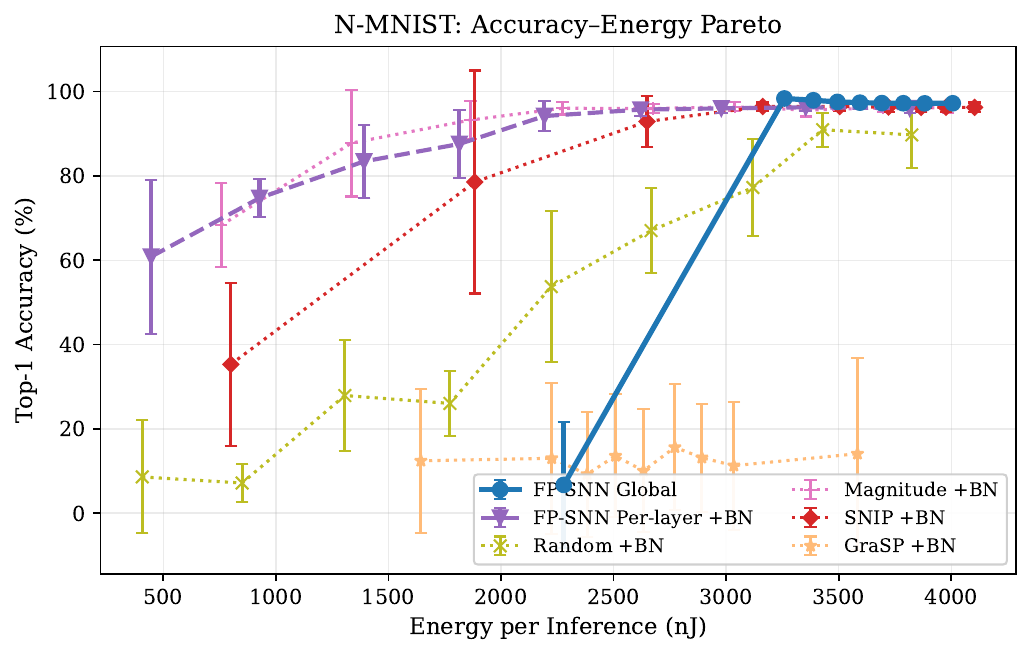}
\caption{Accuracy--energy Pareto frontier on N-MNIST. FP-SNN Global (solid
blue) achieves high accuracy ($\geq 97\%$) only at high energy (dense model),
then dramatically reduces energy at $\sigma=0.8$ while \emph{increasing}
accuracy to $98.4\%$. FP-SNN Per-layer+BN (dashed purple) traces a smoother
frontier, trading accuracy for energy more gradually. All baselines occupy
the lower-left region (low accuracy, low-to-moderate energy), confirming that
they do not achieve a favourable accuracy--energy tradeoff at any operating
point.}
\label{fig:nmnist_energy}
\end{figure}

\section{Discussion}
\label{sec:discussion}

\subsection{What the SNN Setting Reveals About Activation-Energy Pruning}

In the conventional network setting, Fine-Pruning~\citep{BINGHAM2025101242}
justified the activation-energy criterion empirically: it works, and the
biological analogy provides intuition for why.
The SNN setting provides something stronger: it transforms the criterion from
an analogy into a measurement.
Because $s_i(x,t) \in \{0,1\}$, the quantity $E[w_{ij}] = |w_{ij}|\cdot
\sum s_i(x,t)$ is literally the number of AC operations performed through
synapse $(i,j)$ on the target workload, weighted by the connection's magnitude.
On neuromorphic hardware, this is a principled proxy for the energy consumed by
that synapse~\citep{horowitz2014computing,attwell2001energy}: it approximates
but does not exactly equal the true hardware energy, which also depends on
firing rates that change after pruning and on implementation-specific costs.
Pruning by activation energy is therefore optimising a physically grounded
proxy for the quantity we want to minimise -- principled and computable
without gradient computation, but not a claim of exact energy optimisation.

This physical grounding also explains the magnitude of the advantage over
gradient-based baselines.
In a conventional network, SNIP and GraSP estimate saliency through a
differentiable proxy that happens to correlate with importance.
In an SNN, the binary spike function means there is no smooth proxy: the
surrogate gradient approximates the spike-function derivative during training,
but this approximation becomes unreliable as a post-hoc saliency measure when
applied to a fixed network on a narrow target distribution.
Activation energy, by contrast, requires no approximation -- it reads off
directly from the forward pass.

\subsection{The BN Recalibration Crossover}

The crossover in BN recalibration benefit around $\sigma \approx 0.4$
reflects a competition between two effects.
At low sparsity, the activation distribution shifts modestly after pruning,
and the source BN statistics remain close to the true pruned-model statistics.
Re-estimating from 2,000 target samples introduces noise that outweighs the
small distribution shift.

At high sparsity, the shift becomes large -- especially under global scope,
which can nearly eliminate some layers -- and the source statistics are
drastically wrong.
The LIF threshold, which is fixed at $\theta=0.5$ in our experiments,
makes this particularly harmful: a systematic upward or downward shift in
pre-activations can cause entire layers to enter silent or saturated regimes
where every neuron never fires or always fires.
Recalibration corrects this shift even from a small unlabeled sample, because
the shift is systematic (all weights scale by the same factor when 50--80\%
are zeroed) rather than stochastic.

This insight suggests an adaptive recalibration rule: apply BN recalibration
when $\sigma > \sigma^*$, where $\sigma^*$ is estimated from the within-layer
activation statistics before and after pruning.
We leave this adaptive variant to future work.

\subsection{Lottery Tickets in Neuromorphic Data}

The improvement-under-pruning result on N-MNIST ($+1.2\,$pp at $\sigma=0.8$;
Figure~\ref{fig:nmnist_improvement})
extends a finding from the original Fine-Pruning
work~\citep{BINGHAM2025101242} -- where accuracy improvement was observed on
ImageNet-scale models -- to the neuromorphic setting, and is consistent with
a lottery-ticket interpretation in the personalization setting.
Prior work on sparse subnetworks in conventional networks has focused on matching
rather than exceeding the source network.
Our result shows that in the personalization setting -- where a general model is
pruned to a narrow target distribution -- winning tickets can substantively
outperform the source, not merely match it.

The mechanism is intuitive in the SNN context: the source model's spike-train
representation is shared across all 10 classes, and weights that are important
for the 7 non-target classes introduce noise into the representation of the 3
target classes.
Removing these weights allows the remaining connections to express a cleaner
class boundary.
Whether this effect generalises to larger class spaces and more complex datasets
remains an open question addressed only partially by preliminary results on
Tiny-ImageNet (Appendix~\ref{app:tinyimagenet}); full evaluation with a
properly-trained source model is required.

\subsection{Failure Modes and Scope Selection}

The decapitation failure mode of global scope at $\sigma \geq 0.6$ is a
predictable consequence of the energy distribution in feedforward SNNs.
Spike rates decrease with depth as the binary representation becomes more
abstract and sparse.
This means deep-layer weights systematically accumulate less activation energy
than early-layer weights of comparable magnitude, causing global thresholding
to preferentially prune them.
The per-layer scope avoids this by enforcing equal sparsity across all layers.

However, per-layer scope forces the same fraction of pruning in early layers
that have high spike rates and many redundant connections.
These connections are genuinely redundant and can be pruned more aggressively
under global scope without loss.
The crossover at $\sigma \approx 0.65$ marks the sparsity beyond which the
decapitation effect dominates the early-layer redundancy benefit.

\subsection{Limitations}
\label{sec:limitations}

Several limitations constrain the current work.
First, Tiny-ImageNet source models are undertrained due to GPU memory
constraints, yielding source accuracy ($11.7\%$) well below what a
fully-trained ResNet-19 SNN achieves ($\sim$50\%).
These results are reported in Appendix~\ref{app:tinyimagenet} for completeness;
the qualitative pruning patterns are consistent with the two primary benchmarks,
but the absolute numbers cannot be compared to the SNN literature without full
training, and their inclusion in the main evaluation would, as per reviewer
guidance, weaken rather than strengthen the conclusions.
Evaluation on dedicated neuromorphic benchmarks such as
DVS-Gesture~\citep{amir2017low} or CIFAR10-DVS~\citep{li2017cifar10} has
also not been performed; these datasets would test FP-SNN on richer temporal
spike-train structure than N-MNIST provides.

Second, the method is one-shot and does not perform fine-tuning after pruning.
Post-pruning fine-tuning with the unlabeled target stream (e.g., via
self-supervised objectives) could recover additional accuracy, particularly
at extreme sparsity ($\sigma \geq 0.8$).

Third, the BN recalibration crossover threshold $\sigma^* \approx 0.4$ was
identified empirically; a theoretical characterization would be valuable for
practitioners who cannot afford the diagnostic sweep.

Fourth, we evaluate on a single SNN architecture family (VGG-style and SCNN).
Whether the activation-energy saliency signal generalizes to recurrent SNN
architectures~\citep{maass1997networks} or transformer-based
SNNs~\citep{zhou2022spikformer} is not tested here.

\section{Conclusion}
\label{sec:conclusion}

Fine-Pruning~\citep{BINGHAM2025101242} established activation-energy thresholding
as an effective unsupervised personalisation strategy for conventional deep
networks, achieving approximately 70\% sparsity with maintained or improved
accuracy on ResNet-50 and speech models without labeled data or backpropagation.
This paper examined what happens when the same criterion is applied to spiking
neural networks, where the binary spike representation transforms the saliency
metric from an empirical proxy into a physical measurement of synaptic energy
expenditure.

Three findings emerge, each qualitatively distinct from what the conventional
network results would have predicted.

\textbf{Gradient-based methods consistently fail in the tested settings.}
SNIP, GraSP, and magnitude pruning collapse to near-chance accuracy by
$\sigma=0.2$ on both primary benchmarks, across all five seeds and two
fully-trained architectures.
We attribute this to a systematic incompatibility between surrogate-gradient
saliency and the binary spike-train representation in the tested configurations:
the smooth approximation used during training is not informative as a post-hoc
importance measure on a fixed network evaluated on a narrow target distribution.
Whether alternative surrogate functions or calibration procedures would mitigate
this remains open, and we present this as an empirical characterisation rather
than a claim of fundamental incompatibility.
The consistent pattern across two architectures and two datasets makes this a
useful empirical baseline for the design of future SNN compression methods.

\textbf{Activation-energy pruning can improve accuracy through target
specialisation.}
On N-MNIST, removing 80\% of weights while specialising to a 3-class target
raises accuracy from $97.2\pm0.7\%$ to $98.4\pm0.4\%$ (Wilcoxon $p<0.05$,
Cohen's $d\approx2.1$).
This extends the improvement-under-pruning effect
of~\cite{BINGHAM2025101242} to neuromorphic data.
The pattern is consistent with a lottery-ticket interpretation in the
personalization setting, with the biological analogy to experience-dependent
specialisation presented as suggestive rather than mechanistically demonstrated.

\textbf{LIF dynamics produce a sparsity-dependent BN recalibration crossover
in the tested settings.}
The fixed firing threshold makes SNNs acutely sensitive to pruning-induced
activation shift, causing unsupervised BN recalibration to be harmful at low
sparsity and beneficial at high sparsity in our experiments, with a crossover
near $\sigma\approx0.4$.
We present this as an empirical observation in our specific configuration
(two architectures, $N\approx2000$ calibration samples) rather than a
universal property; the crossover point may vary with calibration set size,
architecture depth, and sparsity schedule.
Nonetheless, the pattern is mechanistically plausible -- the fixed LIF
threshold has no analogue in networks with continuous activations -- and
provides a practical guide for when to apply recalibration.

Together these findings demonstrate that activation-energy pruning is not merely
transferable to SNNs but is arguably more principled there, with the metric
transitioning from an empirical proxy to a physical measurement of synaptic
energy expenditure.
They establish directions for follow-on work: a systematic ablation of the
criterion components (magnitude-only vs.\ spike-count-only vs.\ their product)
to directly validate the joint criterion; adaptive recalibration thresholds
that detect the BN crossover automatically; post-pruning refinement using
self-supervised objectives on unlabeled target data; and evaluation of
activation-energy saliency on fully-trained large-scale SNN models where
hardware energy claims can be verified against physical measurements rather
than CMOS estimates.

\section*{Acknowledgements}

The authors thank the Technion for access to computational resources. We thank the ISF for funding.

\section*{Data Availability}

The CIFAR-100 dataset is available at \url{https://www.cs.toronto.edu/~kriz/cifar.html}.
The N-MNIST dataset is available at \url{https://www.garrickorchard.com/datasets/n-mnist}.
Code and trained model checkpoints are available at
\url{https://github.com/JosephBingham/snn_fp}.

\appendix

\section{Magnitude Pruning: Per-layer Retention Diagnostic}
\label{app:magnitude_diagnostic}

\begin{figure}[h]
\centering
\includegraphics[width=\textwidth]{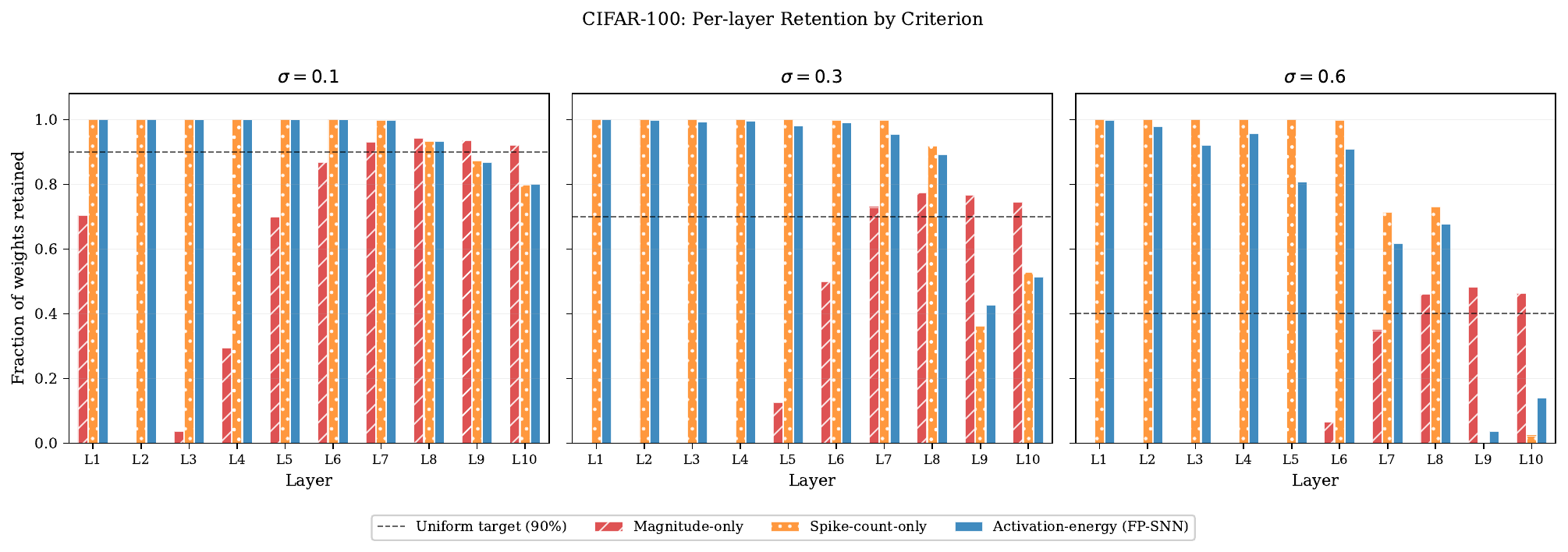}
\caption{Per-layer weight retention for magnitude-only (red), spike-count-only
(orange), and activation-energy / FP-SNN (blue) pruning at $\sigma = 0.1$,
$0.3$, and $0.6$ on CIFAR-100 (seed 0; VGG-SNN, 10 prunable layers L1--L10).
Dashed line: uniform target retention $(1-\sigma)$.
At $\sigma=0.1$, magnitude-only retains nearly 100\% of weights in L1--L3
but near 0\% in L7--L10, decapitating the deep classifier layers while leaving
early feature-detectors intact.
This layer-decapitation pattern explains the immediate accuracy collapse seen
in Table~\ref{tab:ablation_cifar100}: the network loses its task-discriminative
deep representation before losing any early-layer feature capacity.
Spike-count-only and activation-energy pruning distribute retention uniformly
near the target at $\sigma=0.1$; their distributions diverge from each other
only at $\sigma=0.6$, where the weight-magnitude factor in activation-energy
begins to further differentiate among active neurons.}
\label{fig:magnitude_retention}
\end{figure}

Figure~\ref{fig:magnitude_retention} shows per-layer weight retention for all
three criterion-ablation conditions (magnitude-only, spike-count-only, and
activation-energy) at $\sigma = 0.1$, $0.3$, and $0.6$ on CIFAR-100.
This figure addresses the concern that the large performance gap between
magnitude pruning and FP-SNN at low sparsity could reflect an implementation
artefact rather than a genuine saliency-criterion effect.

The retention profiles explain the failure mechanistically.
Under global magnitude thresholding, weights in deep convolutional layers are
systematically smaller than those in early layers -- a direct consequence of
weight-decay regularisation during source training on 100 classes, which
suppresses the magnitude of task-discriminative connections in deep layers
while early, broadly-responding layers retain large weights.
At $\sigma=0.1$, magnitude pruning removes nearly all weights from the deepest
1--2 layers while retaining almost 100\% of early-layer weights.
The network loses its classifier-adjacent representation before losing any
feature-detection capacity, which immediately collapses accuracy to near-chance.

In contrast, activation-energy pruning under global scope distributes retention
much more evenly across layers at low sparsity, because early-layer neurons
that fire broadly across all inputs also accumulate high spike counts -- their
large weights are balanced by high activity, yielding moderate activation-energy
scores and modest pruning.
Deep-layer neurons that fire selectively for target classes accumulate lower
total spike counts but have weights of appropriate magnitude; the product
correctly preserves them.

\section{Preliminary Tiny-ImageNet Results}
\label{app:tinyimagenet}
We report preliminary results on Tiny-ImageNet (200 classes, $64\times64$,
ResNet-19 SNN, 40-class target split, 5 seeds) in this appendix.
Source accuracy is $11.7 \pm 3.1\%$ (chance: 2.5\%), reflecting training
runs cut short at varying epochs due to GPU memory constraints; fully-trained
ResNet-19 SNNs achieve 45--55\%~\citep{fang2021incorporating}.
These results are excluded from the main evaluation because the undertrained
source models do not provide a reliable baseline for assessing pruning
behaviour at the scale this benchmark is intended to test.

Despite this limitation, the qualitative patterns are consistent with the
two primary benchmarks.
FP-SNN Global retains source-level accuracy to $\sigma=0.5$ (10.3\,$\pm$\,3.4\%
vs.\ 11.7\,$\pm$\,3.1\% source) before the global-scope cliff at $\sigma=0.6$
(5.8\%), matching the crossover point observed on CIFAR-100.
FP-SNN Per-layer+BN maintains 9.3\,$\pm$\,2.3\% at $\sigma=0.7$ and
7.9\,$\pm$\,1.4\% at $\sigma=0.8$ -- 3.7$\times$ and 3.2$\times$ above
chance respectively.
Magnitude+BN collapses at $\sigma=0.2$ (3.9\%) and SNIP+BN collapses at
$\sigma=0.4$ (1.4\%), consistent with the patterns on the two primary
benchmarks.

\begin{figure}[h]
\centering
\includegraphics[width=0.75\textwidth]{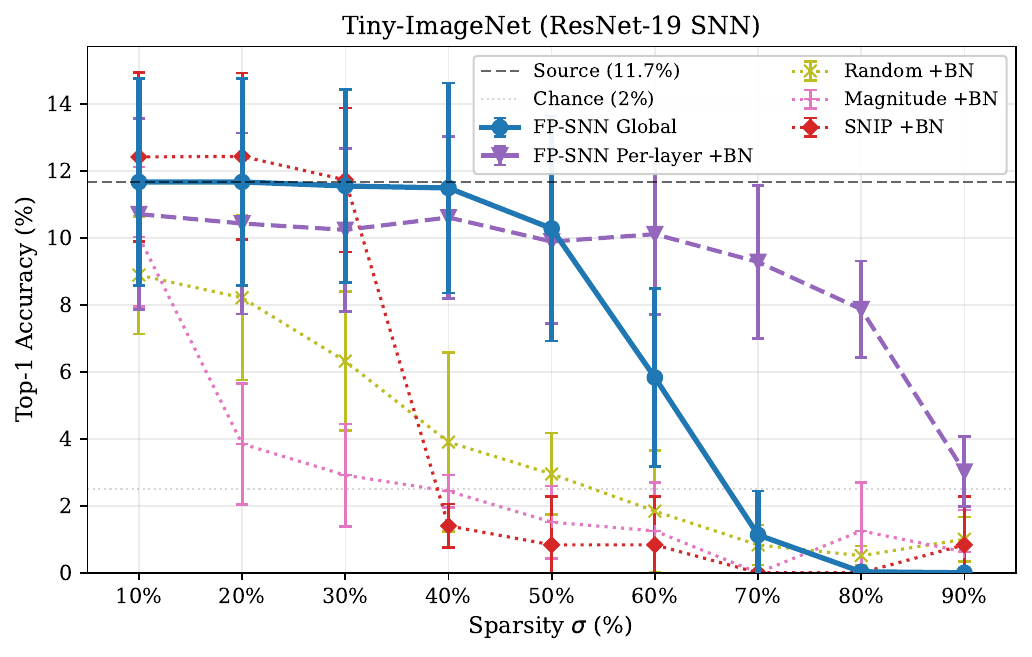}
\caption{Accuracy vs.\ sparsity on Tiny-ImageNet (ResNet-19 SNN, 40-class
target, 5 seeds, undertrained source). All comparisons relative to source.
The qualitative pattern -- FP-SNN Global flat to $\sigma=0.5$, cliff at $\sigma=0.6$,
Per-layer+BN dominant above $\sigma=0.6$, all baselines below chance by $\sigma=0.5$
-- is consistent with the two primary benchmarks.}
\label{fig:tiny_pareto}
\end{figure}

Full Tiny-ImageNet results (Table~\ref{tab:tinyimagenet}) and the Pareto
figure (Figure~\ref{fig:tiny_pareto}) are provided below for completeness.

\begin{table}[h]
\centering
\caption{Tiny-ImageNet preliminary results (ResNet-19 SNN, 40-class target,
5 seeds, undertrained source models at $11.7\pm3.1\%$; chance: 2.5\%).
All comparisons are relative to the $11.7\%$ source, not to literature benchmarks.}
\label{tab:tinyimagenet}
\setlength{\tabcolsep}{4pt}
\begin{tabular}{lcccccc}
\toprule
 & \multicolumn{6}{c}{Sparsity $\sigma$} \\
\cmidrule(lr){2-7}
Method & 0.1 & 0.2 & 0.3 & 0.4 & 0.5 & 0.6 \\
\midrule
FP-SNN global
  & \textbf{11.7\scriptsize{$\pm$3.1}}
  & \textbf{11.7\scriptsize{$\pm$3.1}}
  & \textbf{11.6\scriptsize{$\pm$2.9}}
  & \textbf{11.5\scriptsize{$\pm$3.1}}
  & \textbf{10.3\scriptsize{$\pm$3.4}}
  & 5.8\scriptsize{$\pm$2.7} \\
FP-SNN per-layer+BN
  & 10.7\scriptsize{$\pm$2.8}
  & 10.4\scriptsize{$\pm$2.7}
  & 10.2\scriptsize{$\pm$2.4}
  & 10.6\scriptsize{$\pm$2.4}
  & 9.9\scriptsize{$\pm$2.4}
  & \textbf{10.1\scriptsize{$\pm$2.4}} \\
\midrule
SNIP+BN
  & 12.4\scriptsize{$\pm$2.5}
  & 12.4\scriptsize{$\pm$2.5}
  & 11.7\scriptsize{$\pm$2.2}
  & 1.4\scriptsize{$\pm$0.6}
  & 0.8\scriptsize{$\pm$1.4}
  & 0.8\scriptsize{$\pm$1.4} \\
Magnitude+BN
  & 10.0\scriptsize{$\pm$2.1}
  & 3.9\scriptsize{$\pm$1.8}
  & 2.9\scriptsize{$\pm$1.5}
  & 2.4\scriptsize{$\pm$0.5}
  & 1.5\scriptsize{$\pm$1.1}
  & 1.2\scriptsize{$\pm$1.4} \\
Random+BN
  & 8.9\scriptsize{$\pm$1.8}
  & 8.2\scriptsize{$\pm$2.5}
  & 6.3\scriptsize{$\pm$2.1}
  & 3.9\scriptsize{$\pm$2.7}
  & 3.0\scriptsize{$\pm$1.2}
  & 1.8\scriptsize{$\pm$1.8} \\
\bottomrule
\end{tabular}

\vspace{6pt}
\begin{tabular}{lcccc}
\toprule
 & \multicolumn{4}{c}{High sparsity $\sigma$} \\
\cmidrule(lr){2-5}
Method & 0.7 & 0.8 & 0.9 & Source \\
\midrule
FP-SNN global
  & 1.1\scriptsize{$\pm$1.3} & 0.0\scriptsize{$\pm$0.1} & 0.0\scriptsize{$\pm$0.0} & 11.7\scriptsize{$\pm$3.1} \\
FP-SNN per-layer+BN
  & \textbf{9.3\scriptsize{$\pm$2.3}} & \textbf{7.9\scriptsize{$\pm$1.4}} & \textbf{3.0\scriptsize{$\pm$1.0}} & 11.7\scriptsize{$\pm$3.1} \\
\midrule
SNIP+BN
  & 0.0\scriptsize{$\pm$0.0} & 0.0\scriptsize{$\pm$0.0} & 0.8\scriptsize{$\pm$1.4} & --- \\
Magnitude+BN
  & 0.0\scriptsize{$\pm$0.0} & 1.2\scriptsize{$\pm$1.4} & 0.6\scriptsize{$\pm$1.2} & --- \\
Random+BN
  & 0.8\scriptsize{$\pm$0.6} & 0.5\scriptsize{$\pm$0.3} & 1.0\scriptsize{$\pm$0.7} & --- \\
\bottomrule
\end{tabular}
\end{table}

\bibliographystyle{elsarticle-num}
\bibliography{references}

\end{document}